%% file: iclr2024_conference.tex
\documentclass{article} 
\usepackage[nonatbib, final]{neurips_2023}

\usepackage[numbers, sort, compress]{natbib}

\input{math_commands.tex}

\usepackage{hyperref}
\usepackage{url}
\usepackage{graphicx}
\usepackage{soul}
\usepackage{enumitem}
\usepackage{booktabs}
\PassOptionsToPackage{table}{xcolor}
\usepackage{array}
\usepackage{colortbl}
\usepackage{tabularx}
\usepackage{truncate}
\usepackage{subcaption}
\usepackage{todonotes}

\newcommand{\aaron}[1]{\textcolor{brown}{[Aaron: #1]}}

\newcommand{\captionmethod}{telephoning}
\newcommand{\capcaptionmethod}{Telephoning}
\newcommand{\datasetname}{CommonCatalog}
\newcommand{\modelname}{CommonCanvas}

\newlength{\groupwidth}
\newlength{\itemwidth}

\newcommand{\custompar}[1]{\noindent{\bf #1.}\:}

\newcommand{\prompt}[1]{\scriptsize{\texttt{#1}}}

\definecolor{paleblue}{rgb}{0.678,0.847,0.901}
\definecolor{palepink}{rgb}{1,0.714,0.757}
\definecolor{paleyellow}{rgb}{1,1,0.878}

\usepackage{wrapfig}

\title{\modelname: An Open Diffusion Model Trained with Creative-Commons Images}

\author{
Aaron Gokaslan\textsuperscript{1} \hspace{.15cm}
A. Feder Cooper\textsuperscript{1} \hspace{.15cm}
Jasmine Collins\textsuperscript{2} \hspace{.15cm}
Landan Seguin\textsuperscript{2} \And
Austin Jacobson\textsuperscript{2} \hspace{.15cm}
Mihir Patel\textsuperscript{2} \hspace{.15cm}
Jonathan Frankle\textsuperscript{2} \hspace{.15cm}
Cory Stephenson\textsuperscript{2} \hspace{.15cm}
Volodymyr Kuleshov\textsuperscript{1} \\
\textsuperscript{1}{Cornell Tech} \\
\texttt{\{akg87,afc78,vk379\}@cornell.edu} \\
\textsuperscript{2}Databricks Mosaic \\
\texttt{\{firstname.lastname\}@databricks.com} \\
}

\begin{document}

\maketitle

\input{iclr2023/section/00-abstract}
\input{iclr2023/section/10-intro}
\input{iclr2023/section/20-prelim}
\input{iclr2023/section/30-transfer}
\input{iclr2023/section/40-dataset}

\input{iclr2023/section/50-sys}
\input{iclr2023/section/60-experiments}

\input{iclr2023/section/70-rw}
\input{iclr2023/section/80-acknowledgements}

\bibliography{iclr2024_conference}
\bibliographystyle{plainnat}

\input{iclr2023/section/98-appendix}


\end{document}

%% file: math_commands.tex
\usepackage{amsmath,amsfonts,bm}

\def\eqref#1{equation~\ref{#1}}

\def\1{\bm{1}}

\DeclareMathAlphabet{\mathsfit}{\encodingdefault}{\sfdefault}{m}{sl}
\SetMathAlphabet{\mathsfit}{bold}{\encodingdefault}{\sfdefault}{bx}{n}



%% file: iclr2023/section/00-abstract.tex
\begin{abstract}
We assemble a dataset of Creative-Commons-licensed (CC) images, which we use to train a set of open diffusion models 
that are qualitatively competitive with Stable Diffusion 2 (SD2). 
This task presents two challenges: (1) high-resolution CC images
lack the captions necessary to train text-to-image generative models; (2) CC images are relatively scarce.
In turn, to address these challenges, we 
use an intuitive transfer learning technique 
to produce a set of high-quality synthetic captions paired with curated CC images. 
We then develop a data- and compute-efficient training recipe that
requires as little as 3\% of the LAION data (i.e., roughly 70 million examples) needed to train existing SD2 models, but obtains the same quality. 
These results indicate that we have a sufficient number of CC images (also roughly 70 million) for training high-quality models. 
Our training recipe also implements a variety of optimizations that achieve $\sim$3X training speed-ups,  
and that enable rapid model iteration. 
We leverage this recipe to train several high-quality text-to-image models, which we dub the \emph{\modelname} family.
Our largest model achieves comparable performance to SD2 on human evaluation, even though we only use a CC dataset that is $<$3\% the size of LAION and synthetic captions for training.
We release our models, data, and code at ~\url{https://github.com/mosaicml/diffusion/blob/main/assets/common-canvas.md}.\looseness=-1
\end{abstract}

%% file: iclr2023/section/10-intro.tex
\section{Introduction}\label{sec:intro}

Current methods train high-quality, text-to-image (T2I) models with. 
A lack of curated datasets that are large enough for the task has led researchers to turn to web-scraped solutions~\citep{lee2023explainers, lee2023talkin}, like LAION-2B~\citep{laion2Ben}. 
The use of web-scraped data is a very common practice for training generative models, however, US courts have yet to definitively rule if this is permissible under copyright law~\citep{copilotcomplaint, alphabetcomplaint, getty, kadrey, tremblay, anderson}.  
In response, recent work has begun to investigate alternative methods of navigating copyright concerns in text generation~\citep{min2023silo}, code completion~\citep{copilot-copy-filter,scheffler2022formalizing}, and image generation~\citep{kumari2023ablating}.
Nevertheless, matching the performance of state-of-the-art models remains a challenge. 
In this work, we study the following natural question: \emph{Is it possible to efficiently produce a high-quality T2I model by training only on Creative-Commons-licensed data?} 

We suggest a possible path forward, training 
a suite of T2I architectures using \emph{only} open-licensed, Creative-Commons (CC) images (Figures~\ref{fig:hero-fig} \&~\ref{fig:teaser}). 
This task brings to light two significant challenges. 
The first problem is data incompleteness: almost all CC images lack the captions necessary to train a high-quality T2I model. 
The second is data scarcity: there are relatively few high-resolution CC images --- roughly 70 million, compared to LAION-2B's roughly 2 billion~\citep{laion2Ben}.\looseness=-1 

\input{iclr2023/figs/hero}

We address the data incompleteness problem by using a pre-trained BLIP-2 model~\citep{li2023blip2}, which we use to produce high-quality, synthetic captions for a set of curated, open licensed CC images. 
This is an intuitive transfer-learning solution: leveraging powerful pre-trained generative models to produce synthetic labels for an unlabeled dataset, which we can then use to train a different multimodal generative model. 
We note that this is an increasingly common pattern in the literature, which we shorthand with the name \emph{\captionmethod}. 

To deal with data scarcity, we propose a data- and compute-efficient training recipe that obtains the same quality as SD2, but (perhaps surprisingly) requires as little as 3\% of the LAION-2B data (i.e., roughly 70 million examples) originally used to train SD2. 
We call this model SD2-base. 
These results indicate that we have a sufficient number of CC images (also roughly 70 million) for training high-quality models. 
Our training recipe also implements a variety of optimizations that achieve $\sim$3X training speed-ups,  
and that allow for rapid model iteration. 

The above methods enable us to create \emph{\modelname}, a suite of latent diffusion model (LDM) architectures trained on our curated dataset of CC images and synthetic captions, which we denote \emph{\datasetname}. For CommonCanvasL-NC, we swap SD2's UNet for SDXL to demonstrate how even with less data, larger models do not overfit to this smaller dataset. 
Our largest model achieves performance comparable to SD2-base on human evaluation of Parti Prompts~\citep{yu2022scaling}, even though our \datasetname{} training dataset is $<3\%$ the size of LAION and has synthetically generated captions. 
Figure \ref{fig:hero-fig} shows select samples from our \modelname{}
models compared to corresponding samples from SD2-base. Although this model is a larger and - likely - more capable model architecture than SD2, we find it surprising and important that it is possible to train an SD2-quality model at all based on such a limited dataset that was cobbled together in this fashion. This reveals a promising path forward for future research on highly-capable, open T2I models. In summary, we: 
\begin{itemize}
[topsep=0pt, leftmargin=.5cm, itemsep=0pt]
    \item Synthesize a set of high-quality captions for uncaptioned CC images, which we can then use together for training. We note that this type of transfer-learning technique is increasingly common, and we give it the shorthand name \emph{\captionmethod} (Section~\ref{sec:transfer}). 
    \item Curate \emph{\datasetname}, a dataset of roughly 70 million open-licensed CC images, for which we use \captionmethod{} to generate accompanying high-quality synthetic captions (Section~\ref{sec:dataset}).
    \item Train and evaluate \emph{\modelname{}}, a suite of LDM architectures trained on \datasetname. We demonstrate that these models produce competitive qualitative and quantitative results compared to the 
    SD2-base baseline (Section~\ref{sec:experiments}). To make this analysis tractable, we implement a variety of training optimizations, which achieve $\sim$3X speed-ups in training SD2-base (Section~\ref{sec:mlsys}). 
    \item Release our \datasetname{} dataset of CC images and synthetic captions along with our trained \modelname{} model at \url{https://github.com/mosaicml/diffusion/blob/main/assets/common-canvas.md}.
\end{itemize}

%% file: iclr2023/figs/hero.tex
\begin{figure}[ht!]
    \centering 
    \small
    \setlength{\tabcolsep}{1pt}
    \setlength{\itemwidth}{0.18\linewidth}
    \newcolumntype{M}[1]{>{\centering\arraybackslash}m{#1}}
    \begin{tabular}{M{\itemwidth}M{\itemwidth}M{\itemwidth+.3cm}M{\itemwidth+.55cm}M{\itemwidth+.5cm}}
      Prompt & SD2-base & \modelname-S-C & \modelname-S-NC & \modelname-L-NC \\
    \prompt{a cute black cat inside of a pumpkin} &    \includegraphics[width=\itemwidth]{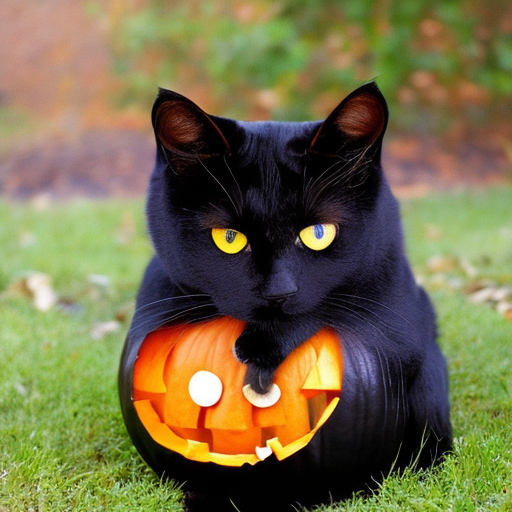}  &
    \includegraphics[width=\itemwidth]{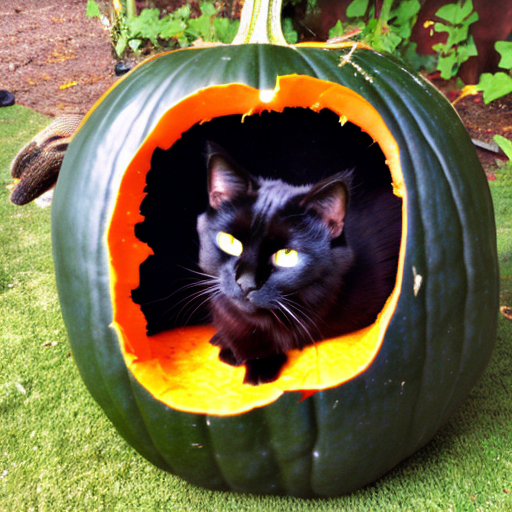} &
    \includegraphics[width=\itemwidth]{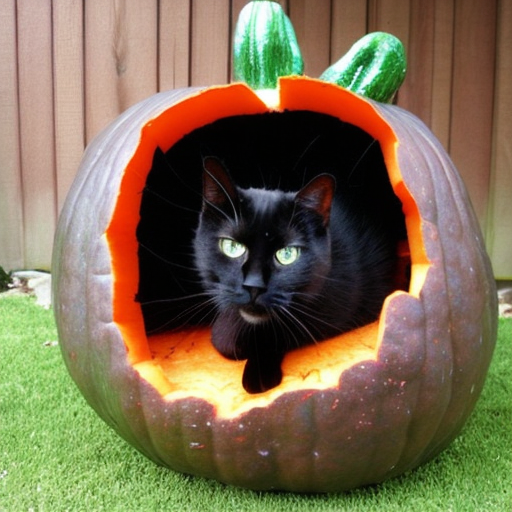} &     
    \includegraphics[width=\itemwidth]{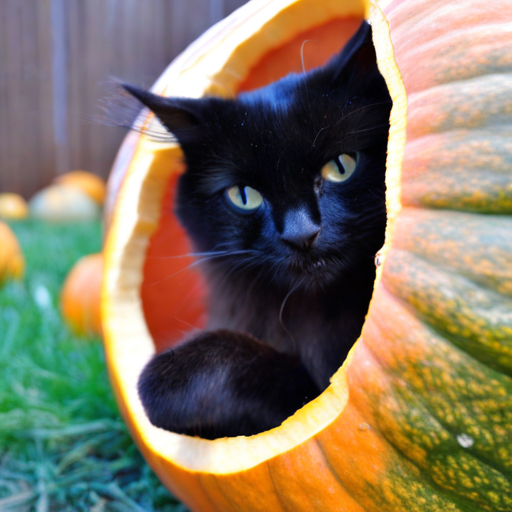} \\

    \prompt{a robot holding a paint palette} &
    \includegraphics[width=\itemwidth]{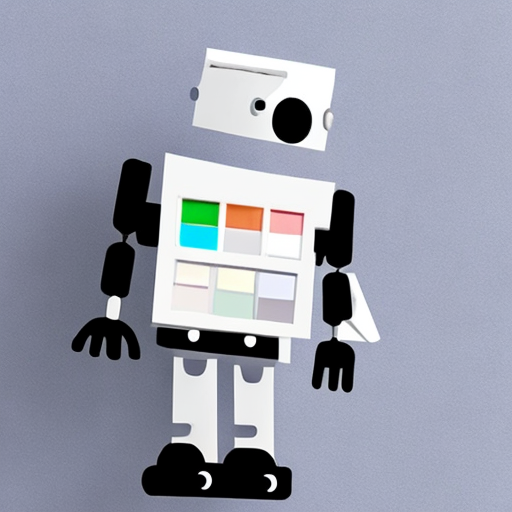}  &
    \includegraphics[width=\itemwidth]{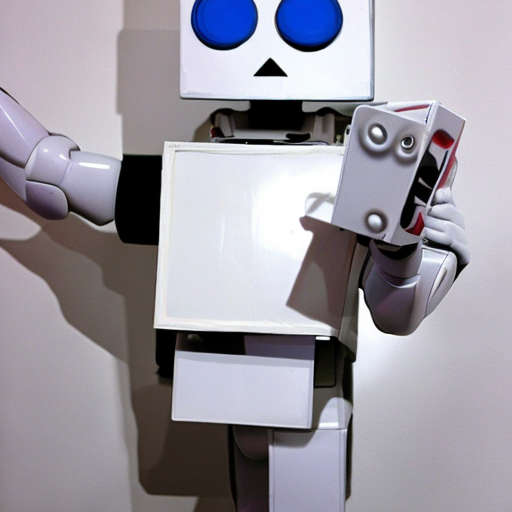} &
    \includegraphics[width=\itemwidth]{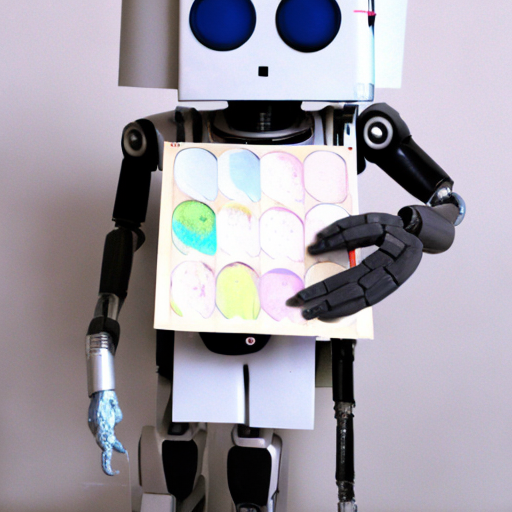} &     
    \includegraphics[width=\itemwidth]{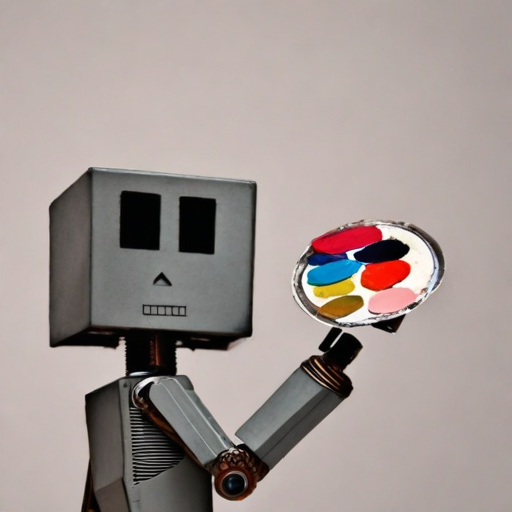} \\

    \prompt{an oil painting of a tall ship sailing through a field of wheat at sunset} &
    \includegraphics[width=\itemwidth]{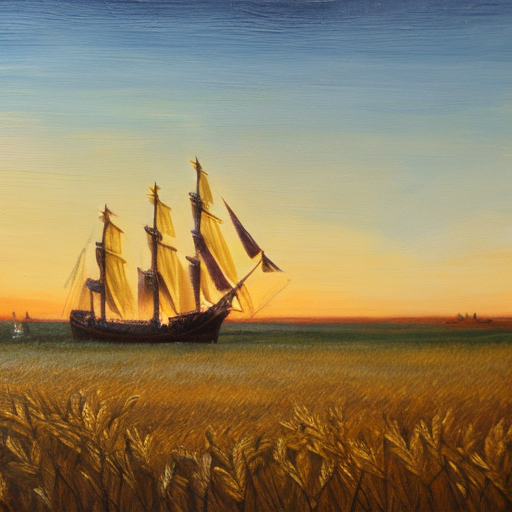}  &
    \includegraphics[width=\itemwidth]{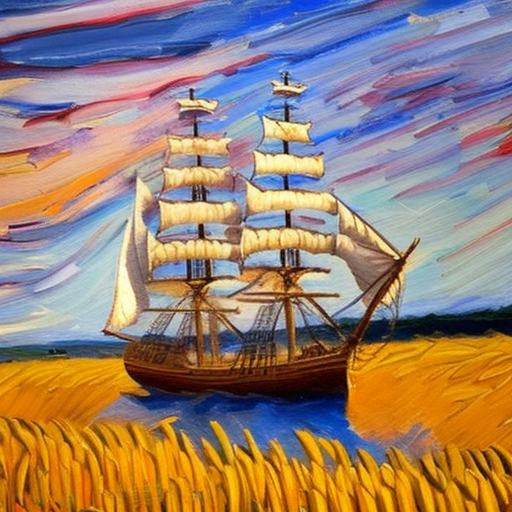} &     
    \includegraphics[width=\itemwidth]{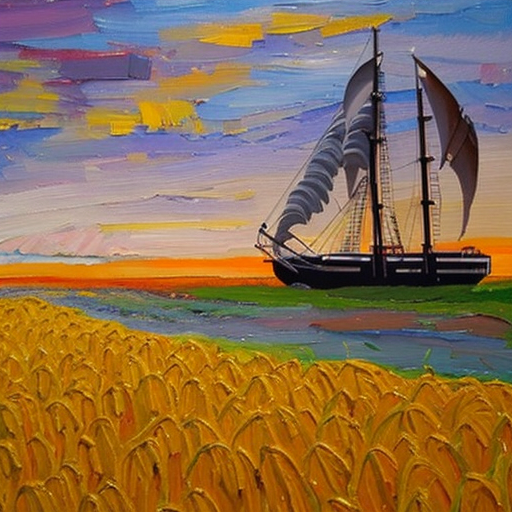} &
    \includegraphics[width=\itemwidth]{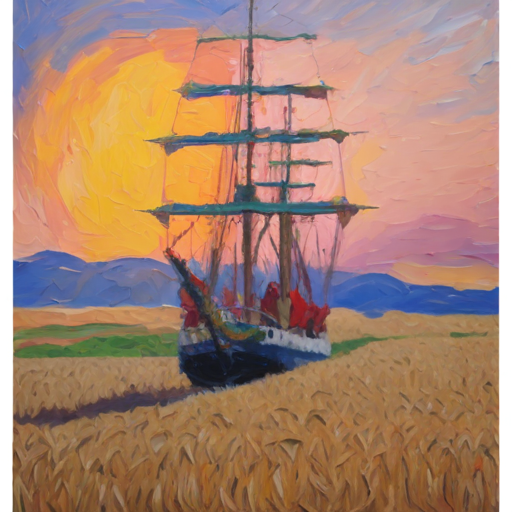} \\
    \end{tabular}
    \caption{Selection of text prompts. Using entirely Creative-Commons images and our synthetic captioning approach, we achieve comparable qualitative performance to Stable Diffusion 2 (SD2-base), as seen in \modelname{} generations, while only requiring a small fraction ($<3\%$) of the amount of training data. We include results for two \modelname{} \textsf{architectures}, small (S) and large (L) (Section~\ref{sec:experiments}), and two CC-image \textsf{datasets}, commercial (C) and non-commercial (NC) (Section~\ref{sec:dataset}). We label our results accordingly as \modelname-$<$\textsf{architecture}$>$-$<$\textsf{dataset}$>$.} 
    \label{fig:hero-fig}
    \vspace{-.5cm}
\end{figure}

%% file: iclr2023/section/20-prelim.tex
\begin{figure}[t]
\centering
\hspace*{-.458cm}
    \begin{minipage}{0.15\linewidth}
        \centering
        \vspace{.42cm}
        \prompt{an image of \\ elsa from \\ frozen}
        \vspace{.65cm}
        \subcaption{Prompt}
    \end{minipage}%
    \hspace{-.035\linewidth}
    \begin{minipage}{0.2\linewidth}
        \centering
        \vspace{-.35cm}
        \includegraphics[width=.78\linewidth]{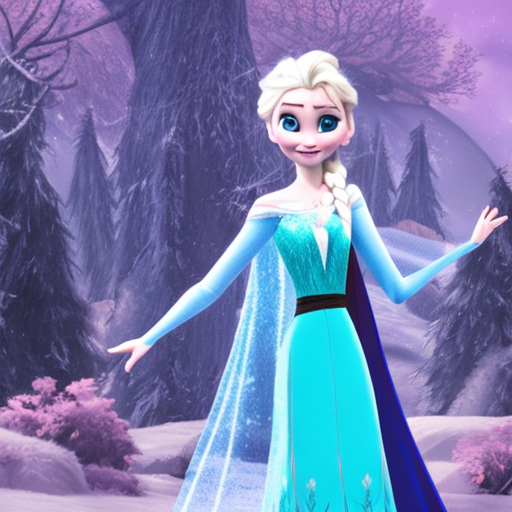}
        \subcaption{SD2 Output}
    \end{minipage}%
    \hspace{-.01\linewidth}
    \begin{minipage}{0.2\linewidth}
        \centering
        \includegraphics[width=.78\linewidth]{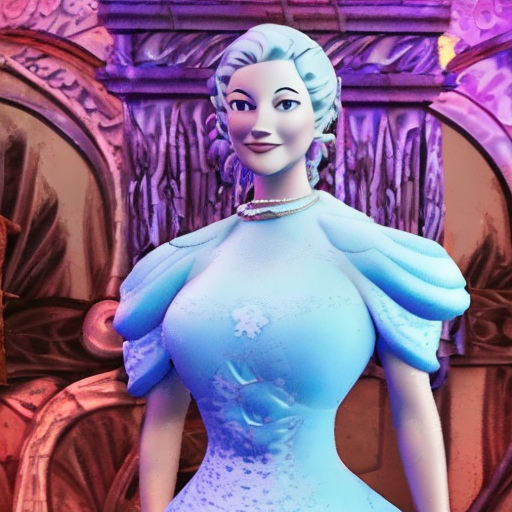}
        \subcaption{\modelname{}\\\hspace*{1cm}Output}
    \end{minipage}
    \hspace{-.02\linewidth}
    \begin{minipage}{0.15\linewidth}
        \centering
        \vspace{.6cm}
        \prompt{the lion king}
        \vspace{1.02cm}
        \subcaption{Prompt}
    \end{minipage}%
    \hspace{-.035\linewidth}
    \begin{minipage}{0.2\linewidth}
        \centering
        \vspace{-.35cm}
        \includegraphics[width=.78\linewidth]{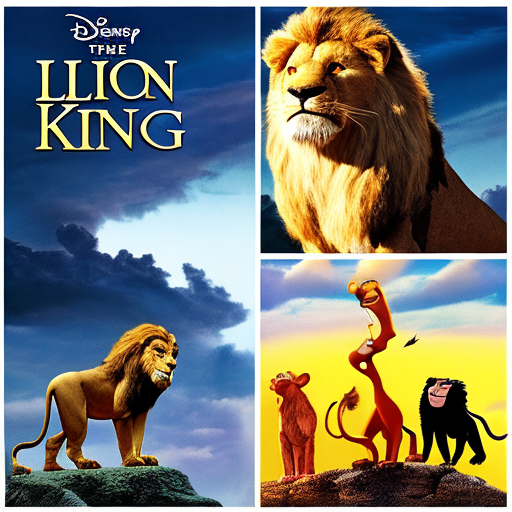}
        \subcaption{SD2 Output}
    \end{minipage}%
    \hspace{-.01\linewidth}
    \begin{minipage}{0.2\linewidth}
        \centering
        \includegraphics[width=.78\linewidth]{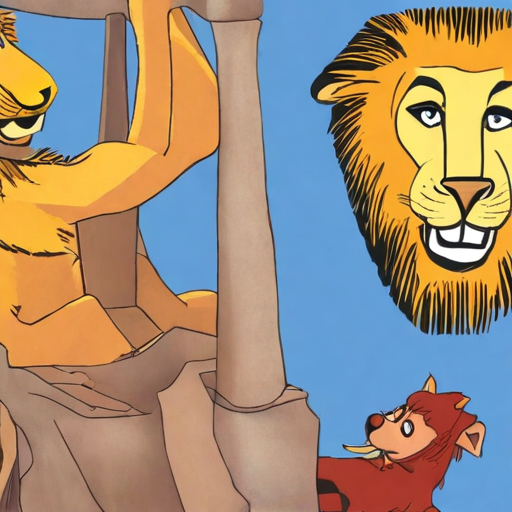}
        \subcaption{\modelname{}\\\hspace*{1cm}Output}
    \end{minipage}
    \caption{When given prompts for concepts related to Disney movies (\textbf{a}, \textbf{d}), SD2-base generates a recognizable image of Elsa from \emph{Frozen}  (\textbf{b}) and a poster-like image with a misshapen Disney logo and characters resembling those from \emph{The Lion King} (\textbf{e}), and \modelname{} (-SC) does not (\textbf{c}, \textbf{f}).}
    \label{fig:teaser}
\end{figure}

\vspace{-.2cm}
\section{Preliminaries and Motivation}\label{sec:prelim}

In this section, we present background on training the T2I Stable Diffusion model, originally trained on the web-scraped LAION-2B dataset. 
We then discuss copyright and reproducibility with respect to  LAION datasets. 
This discussion motivates the creation of an alternative dataset composed of 
open licensed, CC images with synthetic captions, which we introduce in Section~\ref{sec:dataset}.\looseness=-1

\vspace{-.2cm}
\subsection{Text-to-image generative models}\label{sec:diffusion}
\vspace{-.1cm}

Text-to-image (T2I) generative models refer to large neural networks trained on paired image-caption data examples. 
One such family of T2I models is Stable Diffusion (SD)~\citep{rombach2022diffusion}. SD is a latent diffusion model (LDM) that converts images to latent representations and back again using Variational Autoencoders (VAEs)~\citep{kingma2014vae}; 
it uses an iterative sampling procedure ~\citep{sohldickstein2015dpm} and trains an underlying UNet~\citep{ronneberger2015unet}. 
The architecture also includes a text encoder, such as the Contrastive Language-Image Pre-training (CLIP) model~\citep{podell2023sdxl} -- either the original CLIP from OpenAI~\citep{radford2021clip} or its open-source counterpart, OpenCLIP~\citep{cherti2022openclip, ilharco2021openclip}. 

Stable Diffusion 2 (SD2)'s UNet has approximately 865 million trainable parameters; Stable Diffusion XL (SDXL) is larger, with 2.6 billion parameters, and has other advancements involving aspect ratio bucketing, micro-conditioning, and multiple text encoders and tokenizers. 
In terms of training data, the SD-family of models and OpenCLIP are both trained on subsets of the LAION-5B dataset~\citep{laion, schuhmann2022laion}. The exact training dataset for CLIP is unknown, but it is likely webscraped data~\cite{radford2021clip}

\vspace{-.2cm}
\subsection{Copyright and reproducibility in relation to LAION datasets}\label{sec:laion}

LAION-5B is a dataset derived from a snapshot of the Common Crawl, a massive corpus of data scraped from the web. 
From this snapshot, the LAION organization curated pairs of image URLs and their corresponding alt-text captions for the intended use of training T2I and image-to-text (I2T) generative models~\citep{laion, schuhmann2022laion}. 
In practice, T2I models are typically trained on filtered subsets of the full LAION-5B dataset (e.g. LAION-2B~\citep{laion2Ben}). 
Training T2I models on this dataset requires visiting the URLs and downloading the associated images.
There are two elements of LAION datasets that are relevant to our work:\looseness=-1 

\custompar{Copyright} The images associated with LAION datasets have unclear \textit{provenance}:  
it is often not known what the original image sources are~\citep{lee2023explainers, lee2023talkin}. 
Courts have not yet decided if training on these datasets is ``fair use" --- an important exception in copyright~\citep{leval1990toward, sobel2017crisis, lee2023talkin, samuelson}. 
In the interim, there are several copyright lawsuits for the alleged use of LAION-5B subsets to train generative models~\citep{anderson, alphabetcomplaint, getty, gettyverge}. 

\custompar{Reproducibility} Since the datasets only contain the image URLs, and not the images themselves, they are plagued with \textit{link rot}~\citep{lakic2023rot}.\footnote{This also applies to other scraped datasets, such as DataComp~\citep{gadre2023datacomp} and OBELICS~\citep{laurencon2023obelics}.} 
When accessing LAION-5B, there is no guarantee the images still exist at their URLs, making it impossible to fully reproduce the dataset and opening up the possibility of data poisoning attacks~\citep{carlini2023poisoning}. 

A natural alternative 
is to not use LAION datasets for training. One could instead independently curate a dataset of CC-licensed images with known provenance that expressly allow for copying, adaptation, and commercial use. As constituent images can be stored and distributed, this would also solve the link rot problem, thereby enabling greater reproducibility. 
We defer our discussion of sourcing CC-licensed images to Section~\ref{sec:dataset}, where we detail \datasetname: our new, open dataset. 
While CC images are an attractive alternative to LAION-5B, we note that CC images rarely contain the captions necessary to train T2I models. 
Therefore, we first need a method for captioning CC images, which we describe in the next section. 

%% file: iclr2023/section/30-transfer.tex
\begin{figure}[t]
\vspace{-.3cm}
    \centering
    \begin{minipage}{.3\linewidth}
        \includegraphics[width=0.95\textwidth]{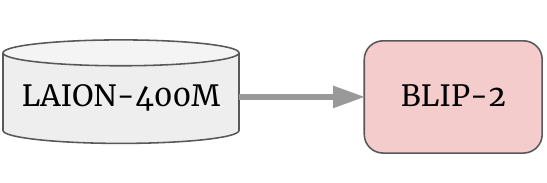}
        \subcaption{Pre-trained BLIP-2.} 
    \end{minipage}
    \hfill
    \begin{minipage}{.6\linewidth}
        \includegraphics[width=0.95\textwidth]{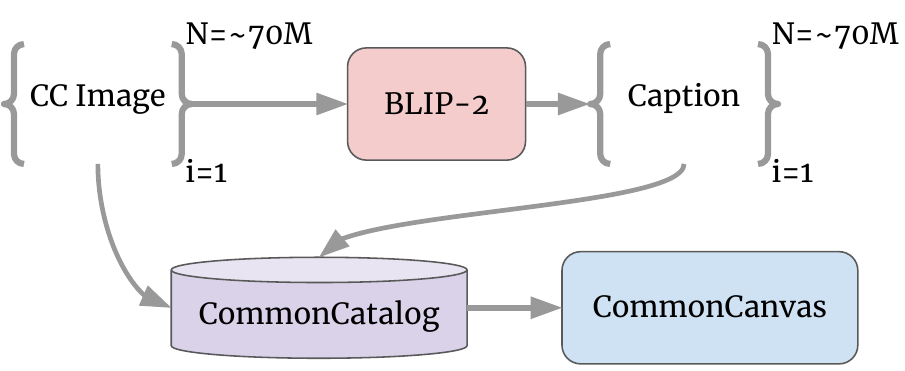} 
        \subcaption{Generating \datasetname{} for training \modelname.}
    \end{minipage}\\
    \begin{minipage}{.95\linewidth}
        \includegraphics[width=0.95\textwidth]{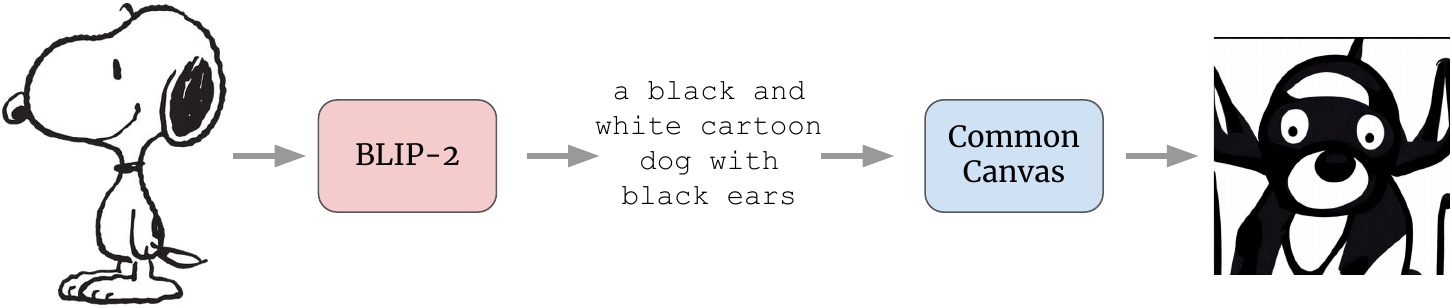} 
        \subcaption{``Lossy compression'' via BLIP-2 from an input image to a synthetic caption. When we use a T2I model to generate an image with this ``lossy'' caption (e.g., via \modelname), the resulting generation looks nothing like the original prompt image that produced the caption.}
    \end{minipage}
    \caption{
    (\textbf{a}) LAION's massive dataset of image-caption pairs is used to 
    train BLIP-2, an image-to-text model. (\textbf{b}) 
    We leverage BLIP-2 to produce synthetic captions for our caption-less 
    CC images, and use the resulting synthetic image-caption pairs (the \emph{\datasetname{}} dataset) to train our open diffusion model, \emph{\modelname{}}. (\textbf{c}) Although BLIP-2 was trained on LAION (e.g., including pictures of characters Snoopy), the captions it produces behave like a ``lossy compression'' (e.g., \texttt{a black and white cartoon dog with black ears}, which has no mention of Snoopy).  When we supply such ``lossy'' captions to a T2I model, like a game of telephone, it produces outputs that 
    no longer resemble the original images (e.g., we show how \modelname{} produces an image that matches the caption, but does not look like Snoopy).}
    \vspace{-.3cm}
    \label{fig:telephoning}
\end{figure}

\section{\capcaptionmethod: A Transfer Learning-based Image-captioning Method} \label{sec:transfer}

Our solution for handling the lack of captions in CC images is an intuitive type of transfer learning for producing high-quality synthetic labels. We describe this method, and then note that there are various similar methods in prior generative modeling literature. Altogether, these methods indicate that this type of transfer learning to produce synthetic labels (to later serve as inputs to training other generative models) has become an increasingly common pattern. We therefore give this method a name: \emph{\captionmethod}. 

\subsection{Describing \captionmethod} 

\capcaptionmethod{} (Figure~\ref{fig:telephoning}) takes inputs from a high-dimensional modality (e.g., images), effectively performs a ``lossy compression'' to a low-dimensional modality (e.g., short-text captions), and then decompresses back to the high-dimensional modality. 
Because the intermediate compression step is ``lossy'', the ultimate output often does not remotely resemble the original input, just like a game of telephone~\citep{telephone}. 
We derive the term \captionmethod{} from the above intuition, and employ it as useful shorthand to denote instances of transfer learning that solve data-scarcity problems in multimodal generative modeling.

In this work, CC images are the high-dimensional inputs, and we use a pre-trained BLIP-2 model~\cite{li2023blip2} for ``lossy compression'' to short-text captions (Figure~\ref{fig:telephoning}a).
Together, these CC-image-caption pairs comprise  the \datasetname{} dataset, which we use to train our \modelname{} T2I models (Figure~\ref{fig:telephoning}b). 
Even though BLIP-2 was pre-trained on LAION-400M~\citep{laion400}, \datasetname{} and \modelname{} never have direct access to LAION-400M or, importantly, anything that is similar to the images that BLIP-2 was trained on. 
Instead, we only have access to the mapping in the model, which, given an image input, produces lossy output text that inherently does not literally resemble its image counterpart (Figure~\ref{fig:telephoning}c).\footnote{We draw on the example of Snoopy from~\cite{sag2023safety}. Figure~\ref{fig:telephoning}'s Snoopy is CC-licensed~\citep{snoopypic}.}   

We defer to experts about fair use (Section~\ref{sec:laion}) --- namely, regarding models like BLIP-2, and LAION-5B's images and alt-text captions. Generally, these experts seem to think that
many cases will fall under fair use~\citep{lee2023talkin, samuelson, lemley2023ai}, 
especially when model outputs do not resemble their inputs, which is the case with BLIP-2. 


\subsection{Related work on \captionmethod}

Our work aligns with the trend of using advanced generative models to address data scarcity.
This is evident in various modalities, such as producing audio captions from image-text pairs~\citep{xiao2023synth} and text from audio~\citep{radford2023robust}. 
Similar approaches have also been used to generate instruction tuning datasets for both text and images ~\citep{li2023self,liu2023visual}. 
Concurrent work has used visual question answers models such as LLava~\cite{liu2023visual} to enhance existing captions such as such as DALLE$\cdot$3~\cite{betker2023improving} and~\citet{chen2023pixartalpha}. However, our model is the one of the first work to train on a dataset without any ground truth captions, and one of the first to release our synthetic captioning dataset along with a fully trained diffusion model. Furthermore, the caption upsampling approaches described in these works could be used to further improve the captions of the CommonCatalogue in future work. Captioning models have been used before to create descriptive captions before to guide a diffusion model to create an image visually similar to a specific image. The concurrent work SynthCap~\cite{caffagni2023synthcap} generates a synthetic captioning dataset using a diffusion model to generate images from captions, tackling the inverse of our problem statement. 

We coin the term \captionmethod{} to shorthand processes like these, which include our work and prior work, and which we believe will become more prevalent as generative models progress.

%% file: iclr2023/section/40-dataset.tex
\section{\datasetname: A Dataset of CC Images \& Synthetic Captions}\label{sec:dataset}

In this section, we introduce our open dataset, \emph{\datasetname}.
First, we describe the collection and curation process for the open-licensed, CC images. 
This process brings to light two challenges: caption-data incompleteness and image-data scarcity. 
To address the lack of CC captions, we show concretely how we use \captionmethod{} to produce high-quality synthetic captions to accompany our set of curated images. 
We investigate the topic of data scarcity in the next section, where we also discuss necessary systems-level training optimizations that enable us efficient SD-model iteration. 

\subsection{Sourcing provenanced, licensed images for \datasetname{}}\label{sec:images}

We focus on locating high-resolution Creative-Commons images that have open licenses. 
We began with the YFCC100M dataset, which consists of 100 million CC-licensed images and multimedia files, as well as Flickr IDs linking to the original data~\citep{thomee2016yfcc100m}. The images in the dataset associated with the original paper exhibit two issues that make it ill-suited for direct use to train Stable Diffusion: they are low-resolution, and many of them have licenses that do not expressly allow for the distribution of derivative works, which are an area of unsettled copyright law in the context of model training. 
We therefore re-scraped these images from Flickr, based on the IDs provided in the YFCC100M metadata. 
Our scraped images are very high resolution (exceeding 4K), which makes them more suitable for T2I training. \vspace{.2cm}
\begin{wrapfigure}{r}{0.51\textwidth}
\vspace{-.5cm}
   \centering
   \caption{\datasetname-C contains images licensed only for commercial use; -NC contains -C as well as images licensed for non-commercial use.\looseness=-1} \vspace{.1cm}
   \label{tab:catalog}
   \footnotesize
    \begin{tabular}{lrr}
    \toprule
        \textbf{Dataset} & \textbf{\# Images} & \textbf{\% Alt Text} \\\midrule
        \datasetname-C & 26,232,417 & 30.76\% 
        \\\midrule
        \datasetname-NC & 67,015,331 & 
        31.22\% 
        \\\bottomrule
    \end{tabular}
\end{wrapfigure}
We exclude images with non-derivative (ND) licenses. 
The remaining images can be further divided into 
those that can be used for commercial (C) purposes and those that cannot (non-commercial/ NC). As shown in Table~\ref{tab:catalog}, we accordingly construct two datasets, 
\datasetname-C and \datasetname-NC. We defer additional details about licenses to  
Appendix~\ref{app:sec:catalog:images}, but emphasize that all of the images included have open licenses: individuals are free to use, adapt, and remix the images, so long as they attribute them. 
In total, \datasetname{} contains roughly 70 million NC CC-images, of which a subset of approximately 25 million images can also be used commercially.\looseness=-1 

Directly sourcing \datasetname{} avoids some concerns (Section~\ref{sec:laion}); 
however, it also comes with its own challenges. 
For one, CC images rarely have the alt-text captions necessary to train a T2I model like Stable Diffusion (Figure~\ref{tab:catalog});
those that do have associated text often just include the image title or a URL.
For another, we could \emph{only} find roughly 70 million usable CC images, which pales in comparison to the billions of images in LAION used to train SD2 (Section~\ref{sec:mlsys}). We take each of these challenges in turn. First, in the next subsection, we show how we instantiate \captionmethod{} (Section~\ref{sec:transfer}) to produce high-quality, synthetic captions for CC images. 

\subsection{Synthesizing captions with \captionmethod{}}\label{sec:captions}

\begin{figure}[t]
  \begin{minipage}{0.15\linewidth}
    \includegraphics[width=\linewidth]{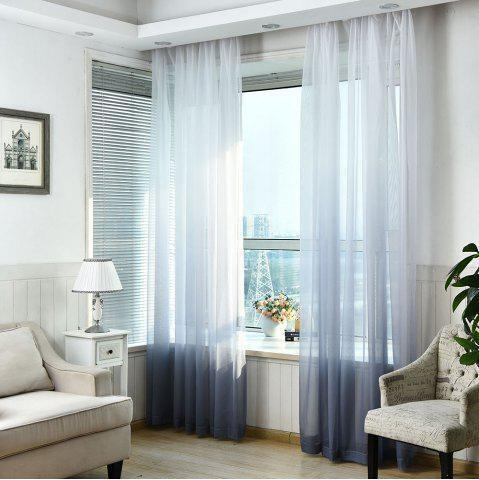}
  \end{minipage}%
  \hfill
  \begin{minipage}{0.83\linewidth}
\begin{tabular}{p{0.3\linewidth}p{0.6\linewidth} }
    \toprule
    \textbf{Source} & \textbf{Caption} \\\midrule
    Alt-Text (LAION-2B) & \prompt{Latest 1PC Transparent Gradient Color Voile} \prompt{Window Curtain} \\
    BLIP2-OPT-2.7B & \prompt{A living room with a white couch and curtains} \\
\end{tabular}
  \end{minipage}
  \caption{Original vs. BLIP-2-generated captions for an image from LAION-2B. 
  BLIP-2 generates a caption that better aligns with what a human would  write. See Figure~\ref{fig:blip-roundtrip} for more examples.}
  \label{fig:blip2-example-caption}
  \vspace{-.5cm}
\end{figure}

We compared several captioning models and, based on qualitative analysis and its state-of-the-art performance on MS COCO, chose to use the pre-trained BLIP-2 OPT2.5B model for synthesizing \datasetname's captions~\citep{li2023blip2}. 
BLIP-2 consists of three components: a pre-trained, frozen (i.e., fixed) visual encoder, a learned transformer network that converts the visual embeddings into a text prompt, and a frozen large language model (LLM) that takes in the prompt. 
The only trainable variables in the transformers are between the frozen visual encoder and frozen LLM layers. 

Given a LAION-2B image as input, we found that the resulting BLIP-2 caption is often qualitatively more descriptive than the corresponding LAION-2B ground-truth alt-text caption.
LAION-2B captions often contain product names, irrelevant details, or poor grammar and syntax (Figure~\ref{fig:blip2-example-caption}).
This finding is corroborated by \citet{nguyen2023improving}, which shows quantitatively (in terms of CLIP Score) that BLIP-2 captions are higher quality than ground-truth captions, at the cost of caption diversity. 

Based on these preliminary results, we captioned all of the YFCC100M Creative-Commons images, which required about 1,120 GPU A100 hours. 
To do so, we center-cropped and resized all of the images to a maximum size of 512x512 pixels. We perform these transformations because captioning images at native resolution would be very expensive. At training time of the diffusion model, all images remain in their native resolution.
We release our commercial (\datasetname-C) and non-commercial (\datasetname-NC) CC-image and synthetic-caption datasets on HuggingFace at [REDACTED] with associated data cards. 
As an evaluation set, we also release the BLIP-2 captions that we produced for the non-derivative (ND) CC images that we did not use for training.

%% file: iclr2023/section/50-sys.tex
\section{Training Efficiency Optimizations and Data Scarcity Analysis}\label{sec:mlsys}
\vspace{-.2cm}

High-resolution CC images are indeed much less abundant than arbitrary web-scraped ones, but the amount of data necessary to train high-quality SD2 models has not been well-studied. 
We set out to quantify this amount by training multiple SD2 models on differently-sized subsets of LAION-2B. 
However, training a single SD2 model, even with hundreds of GPUs, can take several days. To make our data scarcity analysis more tractable, we first implement several efficiency optimizations. 

\vspace{-.1cm}
\subsection{Software and hardware speed-ups}\label{sec:speed}
\vspace{-.1cm}



\begin{figure}[b]
\vspace{-.4cm}
    \centering
    \setlength{\itemwidth}{0.45\linewidth}
    \begin{minipage}{\itemwidth}
        \centering
        \includegraphics[width=\itemwidth]{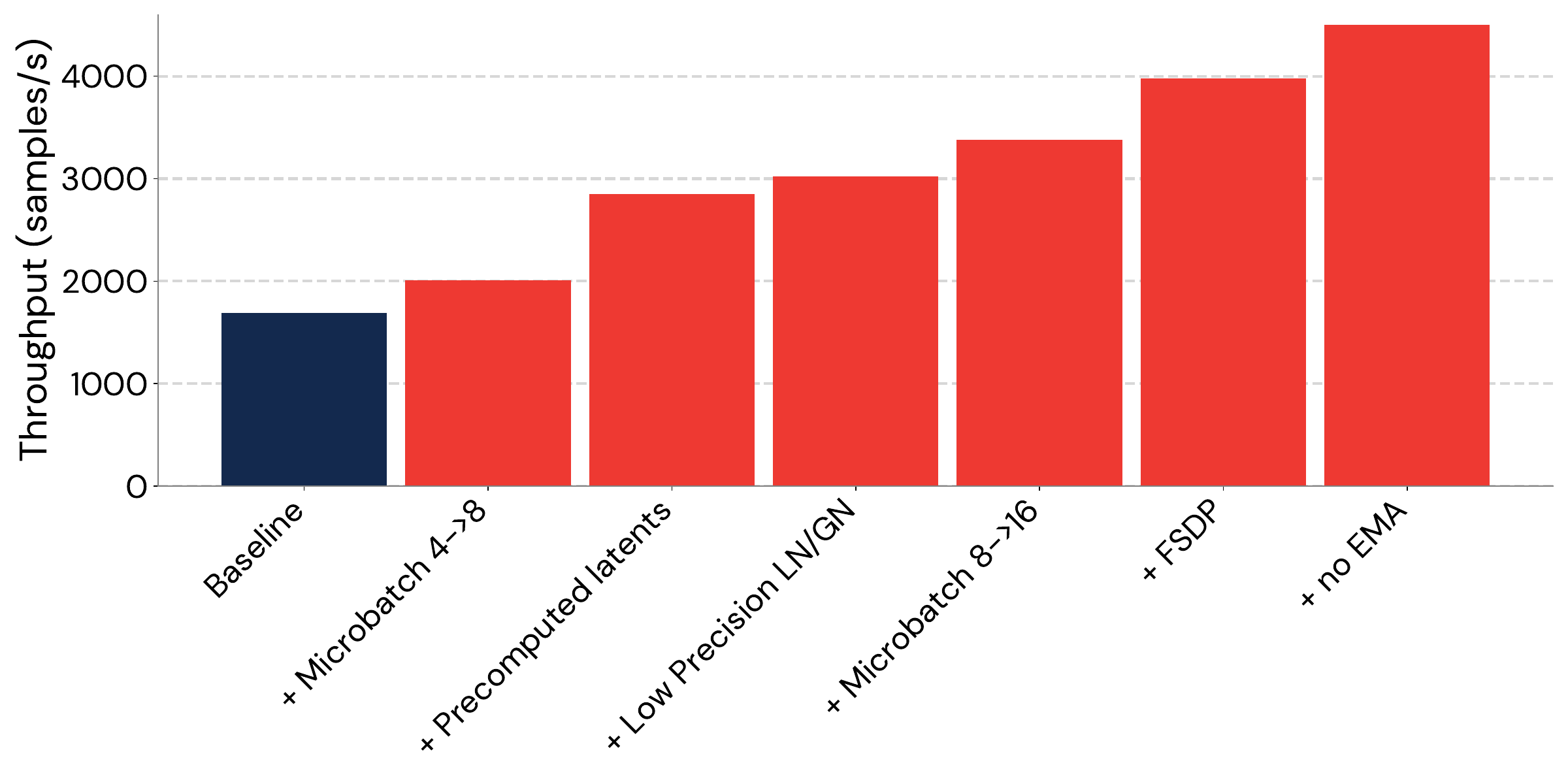}
        \caption{Cumulative effect of various speed-ups in our SD2 training pipeline on 128 Throughputs evaluated on 128 A100s.
        }
        \label{fig:benchmark-speedup}
    \end{minipage}\hspace{1cm}%
    \begin{minipage}{\itemwidth}
        \centering
        \includegraphics[width=\itemwidth]{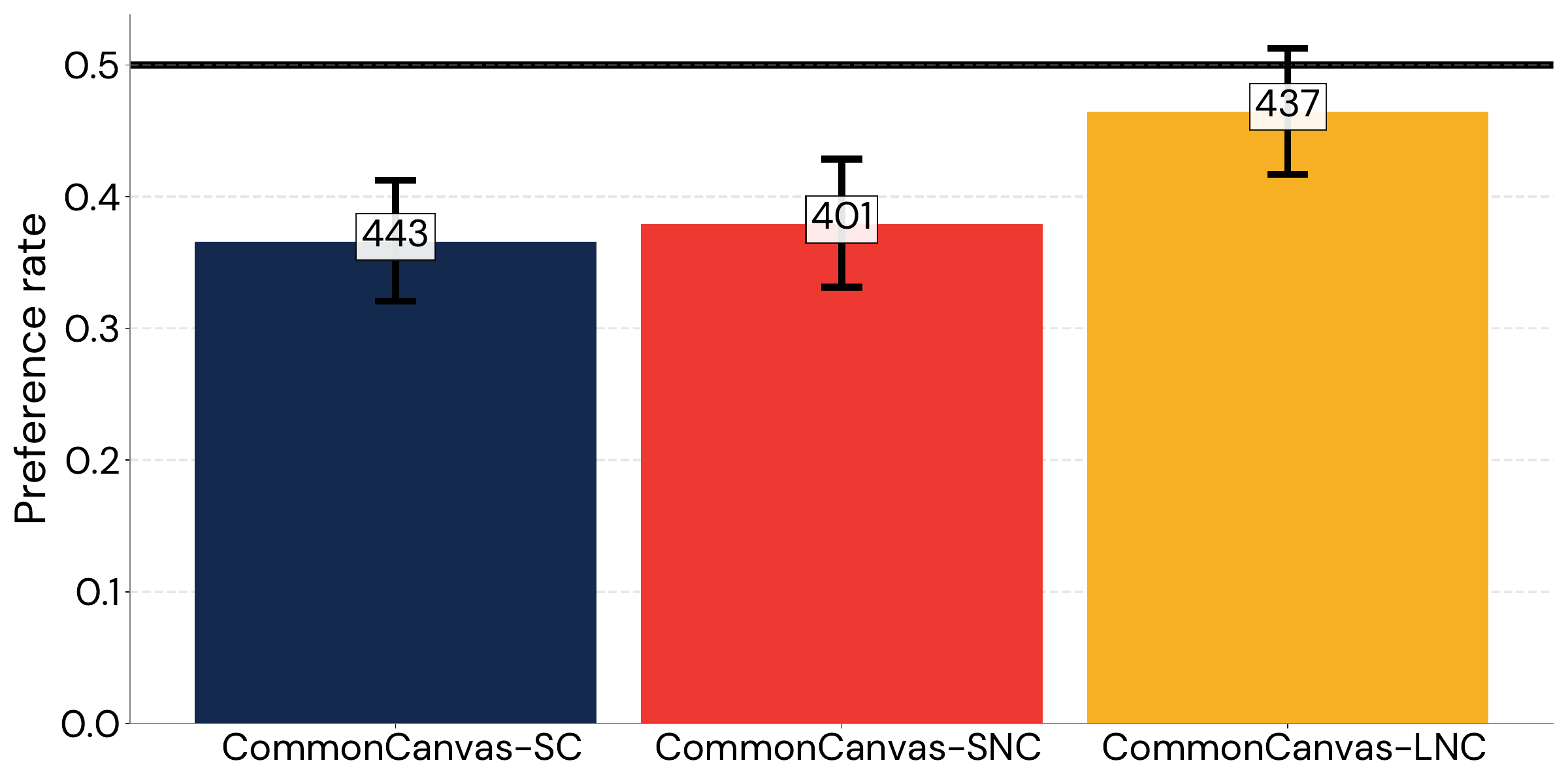}
        \caption{User preference study using Parti prompts.
        \modelname{}-LNC model matches the performance of SD2 despite being trained with $<3\%$ the amount of data.}
    \label{fig:parti-prompts} 
    \end{minipage}.
\end{figure}

Stability AI reports an estimated 200,000 A100 hours to train SD2~\citep{sd2modelcard}. 
Depending on the available hardware, a single SD2 run could take anywhere from a few weeks to over a month to train. 
We sought out multiple avenues to reduce this training-time constraint. Ultimately we were able to achieve a speedup of 2.71X relative to the original SD2 implementation. 


First, we applied Flash Attention~\citep{dao2022flashattention} with the xFormers library~\citep{xFormers2022}. We also pre-computed VAE and text encoder latents over the entire training dataset, cast all GroupNorm~\cite{wu2018group} and LayerNorm~\cite{ba2016layer} to \textsf{float16} precision, and applied fully-sharded data parallelism (FSDP) to our training run. Finally we opted to only keep an exponential moving average of the weights for the final 3.5\% of training. More detail on each of these improvements can be found in Appendix~\ref{app:sec;mlsys}. 

When applying all of the aforementioned strategies together, we are able to achieve a 2.71X speedup in A100 hours over our SD2-baseline implementation. We found that latent pre-computation helped the most at low resolutions, while FSDP also provided significant gains, especially at scale. 
The other optimizations helped reduce total memory usage, allowing us to increase the microbatch size for better hardware utilization. Figure \ref{fig:benchmark-speedup} summarizes each of the proposed methods and the cumulative speedup that results from its application. Equipped with an optimized training setup, we are able to more
easily study effect of varying training dataset size.

\vspace{-.1cm}
\subsection{Investigating data scarcity: Saturating SD2 evaluations with $<3\%$ of LAION-2B}\label{sec:descale}
\vspace{-.1cm}

YFCC100M contains 100 million images, about 10\% the size of the 1.1B LAION examples we could access, thus about 5\% of the original LAION-2B dataset. 
One interesting question that remains unanswered is how much data is actually needed to train these diffusion models effectively. 

\begin{figure}
\centering
\includegraphics[width=\textwidth]{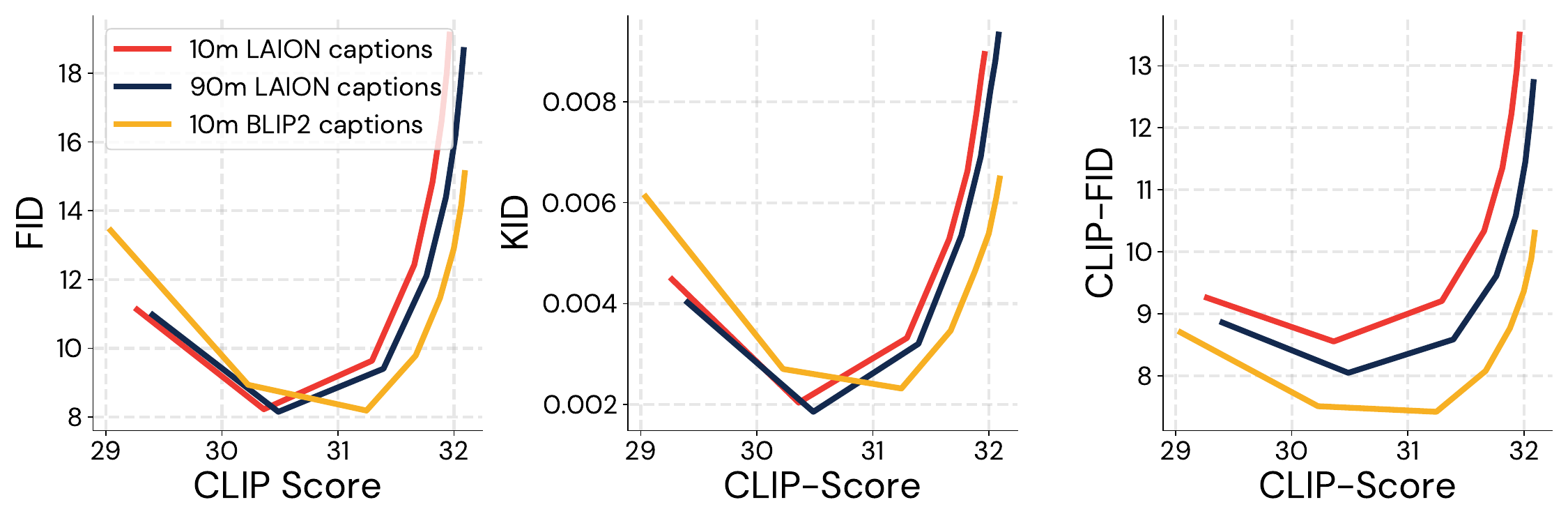}
\caption{FID, KID, and CLIP-FID vs. CLIP-Score computed on 30K samples from COCO2014 for different SD2 models trained on smaller subsets of LAION (10M, 90M, using either original captions or synthetic BLIP2 captions. Interestingly, increasing the amount of training data from 10M to 90M samples does not lead to improved quantitative metrics across guidance scales 1 to 8. Lower FID is better; higher CLIP score is better.}
\label{fig:data-variance}
\end{figure}

We ask whether or not it is necessary to train on 1+ billion images to get results that are as good as the original LAION-trained SD2. Our results show, surprisingly, that this is not the case with a slightly larger model (CommonCanvas-L); this model replaces SD2's U-Net with SDXL's~\citep{podell2023sdxl} larger one. Further, our larger model achieves comparable results to SD2-base on human evaluation, using 33X less training data. We train on increasingly smaller, random subsets of data from our LAION-1.1B model and find that we can achieve a similar result on the commonly reported MS COCO numbers, but with $<$3\% the amount of SD2's training data (Figure~\ref{fig:data-variance}). In fact, we run experiments down to 1-million LAION-1.1B images, and find that only 10 million images are required for stable training behavior (Appendix, Figure~\ref{fig:eval-over-time-less-data}). 

\subsection{Investigating the performance of CC trained model}

These findings suggest that SD2 models may be underparameterized. In fact, when we use CommonCanvas-LNC, we achieve competitive performance with SD2 on user preferences, despite training on significantly less data (Section~\ref{fig:parti-prompts}). Further, in spite of the drastic reduction in dataset size, we observe that the larger model (CommonCanvas-LNC) outperforms the smaller one (CommonCanvas-SNC), consistent with the notion that these models are still underparameterized. We hypothesize about why this might be the case and how much data is actually necessary to saturate the model in Appendix~\ref{app:sec;diff-too-small}.

%% file: iclr2023/section/60-experiments.tex
\vspace{-.2cm}
\section{Experiments }\label{sec:experiments}
\vspace{-.2cm}

\input{iclr2023/wandb_charts/conc-caps-subsets/concept-captions-fig}

Equipped with commercial (\datasetname-C) and non-commercial (\datasetname-NC) datasets, we train two different \modelname{} models. 
We additionally train a larger variant of \modelname-NC (\modelname-LNC) that, as we note above (Section~\ref{sec:descale}), has a significantly larger U-Net. Figure~\ref{fig:hero-fig} displays qualitative results from each of these model variants. More details on the \modelname-LNC architecture can be found in Appendix~\ref{app:sec;largeboi}.

\subsection{Automated quality metrics for model evaluation}


We measure performance with three automated image quality metrics on the MS COCO dataset~\citep{lin2014microsoft}: Frechet Inception Distance (FID)~\citep{hessel2021clipscore}, Kernal Inception Distance (KID)~\citep{binkowski2018demystifying}, and CLIP-FID~\citep{kynkaanniemi2022role}. Additionally, CLIP Score was evaluated to understand the alignment between captions and their respective images. Our model demonstrated comparable performance compared to the baseline of SD2 on the popular MS COCO benchmark.

However, like any model, ours has limitations. It underperformed in several categories, including faces, general photography, and paintings. These categories originated from the Conceptual Captions dataset~\citep{sharma2018conceptual}, which relies on web-scraped data. These web-sourced captions, while abundant, may not always align with human-generated language nuances. 

This discrepancy underscores the importance of incorporating large-scale, human-generated caption data. Although transitioning to synthetic captions introduces certain performance challenges, the drop in performance is not as dramatic as one might assume. Moreover, we speculate that it would if users were to supplement with their own datasets, like FFHQ~\citep{karras2019style}, if they seek to fine-tune models for specific categories.

\begin{figure}[t]
    \centering
    \setlength{\groupwidth}{0.31\linewidth}
    \setlength{\itemwidth}{0.5
    \groupwidth}
    \setlength{\tabcolsep}{0pt}
    \newcolumntype{C}[1]{>{\centering\arraybackslash}p{#1}}
    
    \begin{tabular}{cc@{\hskip 0.05in}cc@{\hskip 0.05in}cc}
    
    Ours & SD2 & Ours & SD2  & Ours & SD2 \\
    
    \includegraphics[width=\itemwidth]{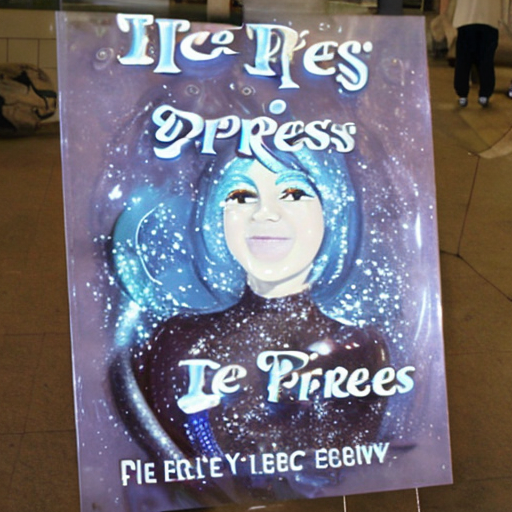} &
    \includegraphics[width=\itemwidth]{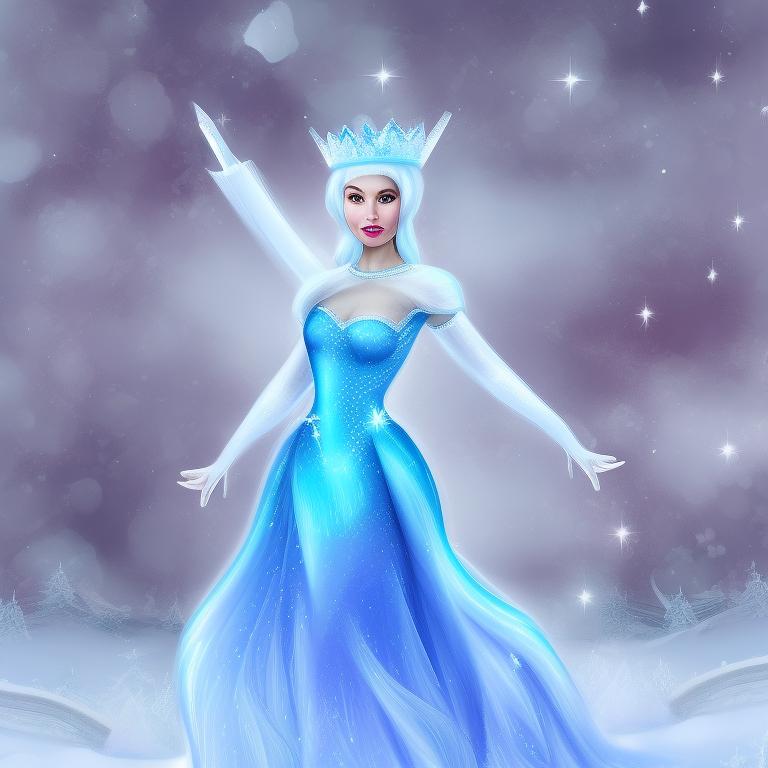} &
    
    \includegraphics[width=\itemwidth] {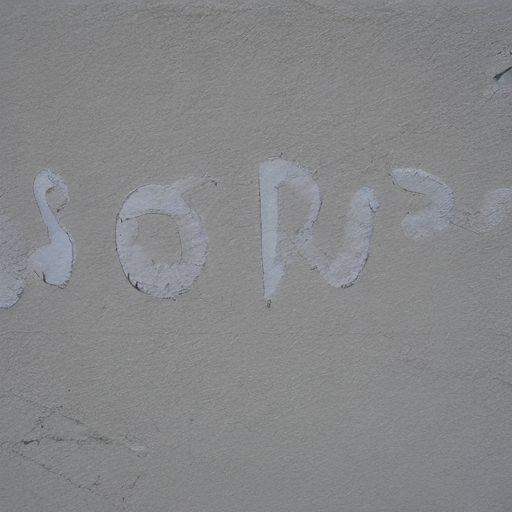} &
    \includegraphics[width=\itemwidth]{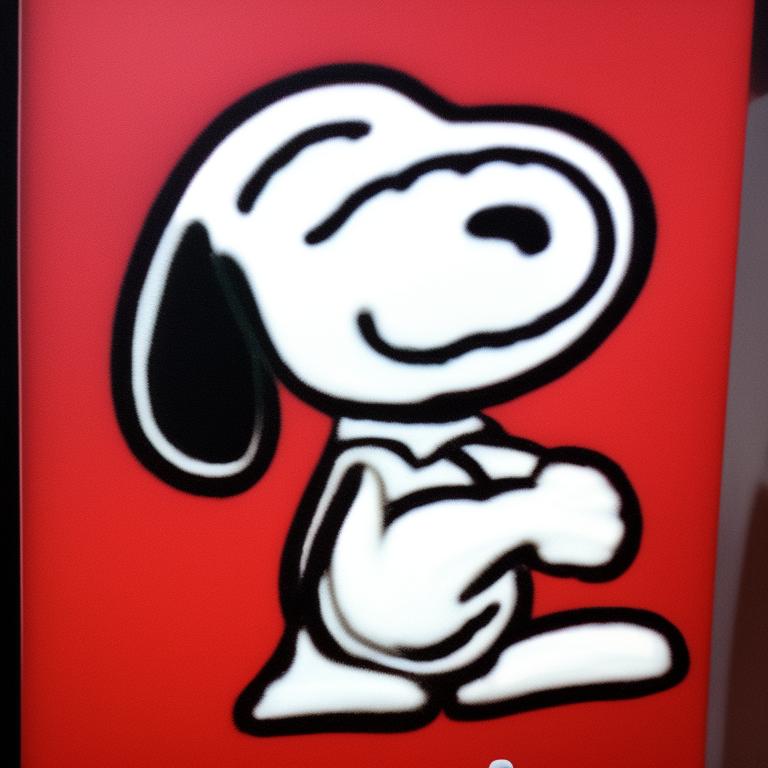} &
    
    \includegraphics[width=\itemwidth]{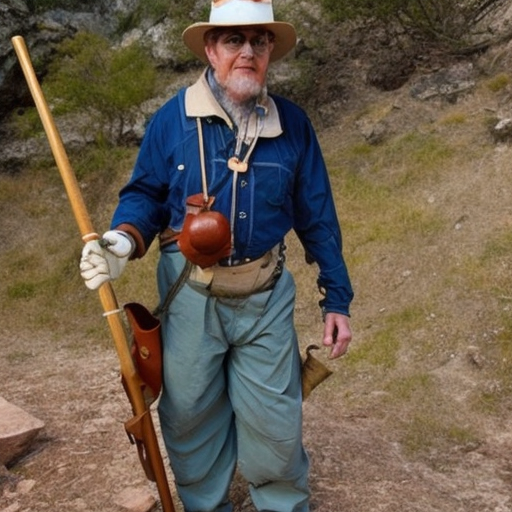} & %
    \includegraphics[width=\itemwidth]{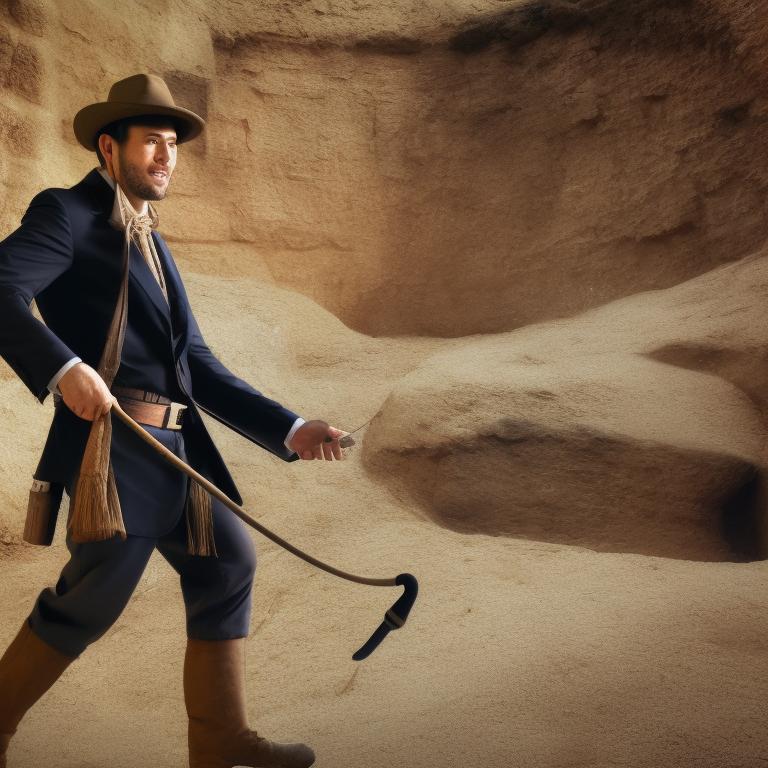} \\
    
    \multicolumn{2}{C{\groupwidth}}{\prompt{ice princess}} &
    \multicolumn{2}{C{\groupwidth}}{\prompt{Snoopy}} &
    \multicolumn{2}{C{\groupwidth}}{\prompt{a adventurous archaeologist with a whip and a fedora}} \\
    
    \includegraphics[width=\itemwidth]{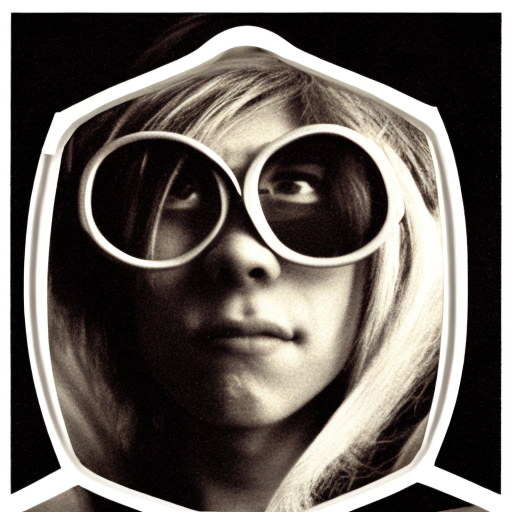} &
    \includegraphics[width=\itemwidth]{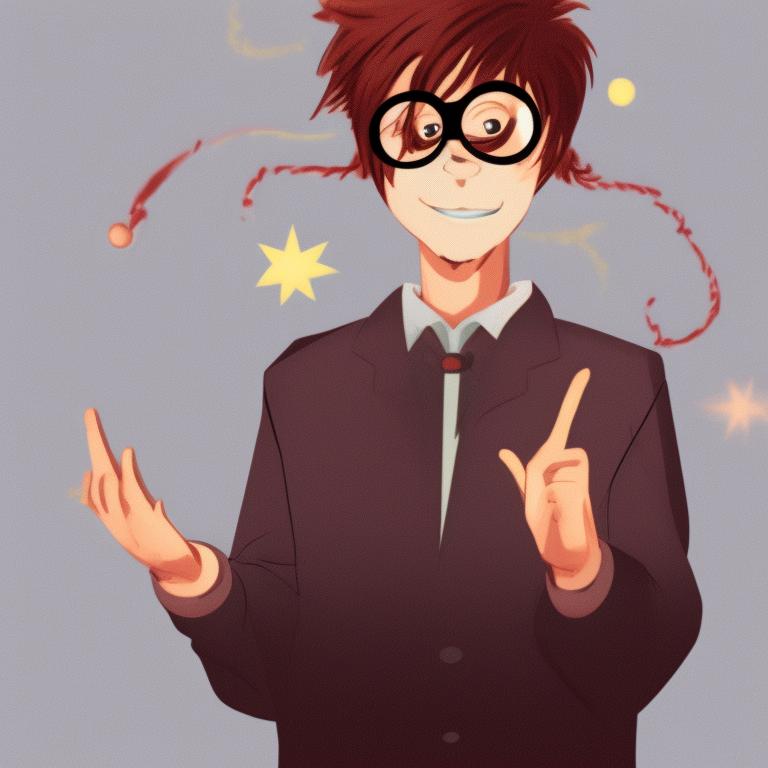} &
    
    \includegraphics[width=\itemwidth]{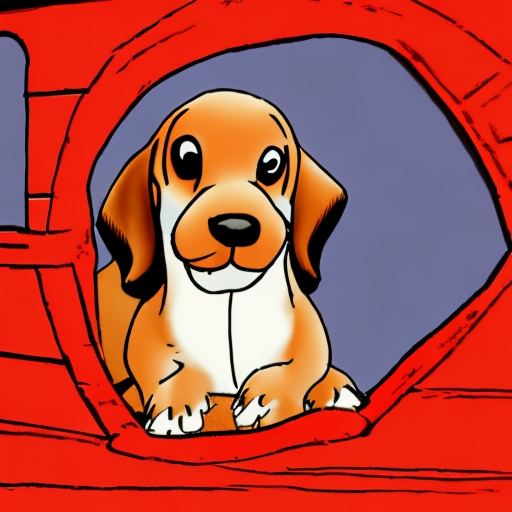} &
    \includegraphics[width=\itemwidth]{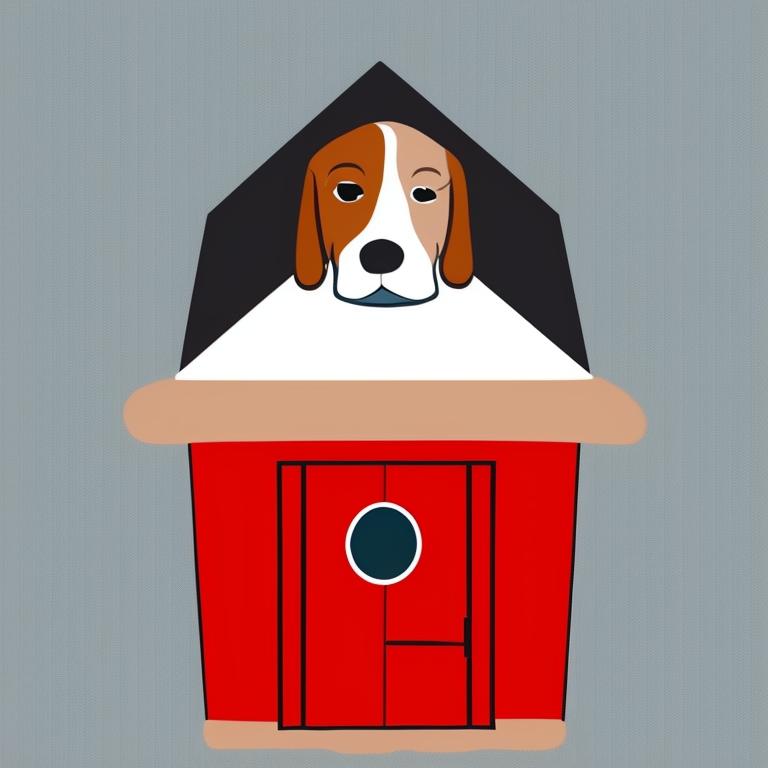} &
    
    \includegraphics[width=\itemwidth]{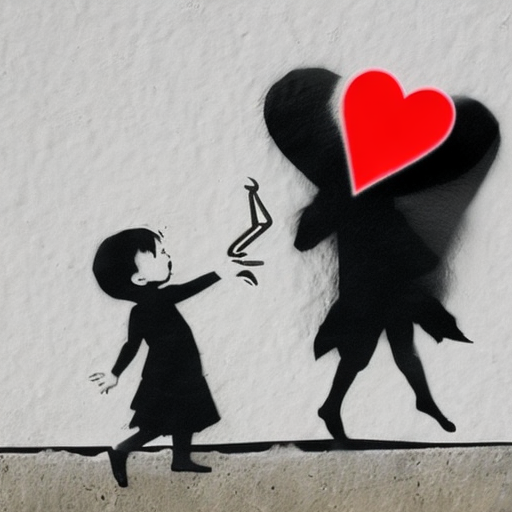} &
    \includegraphics[width=\itemwidth]{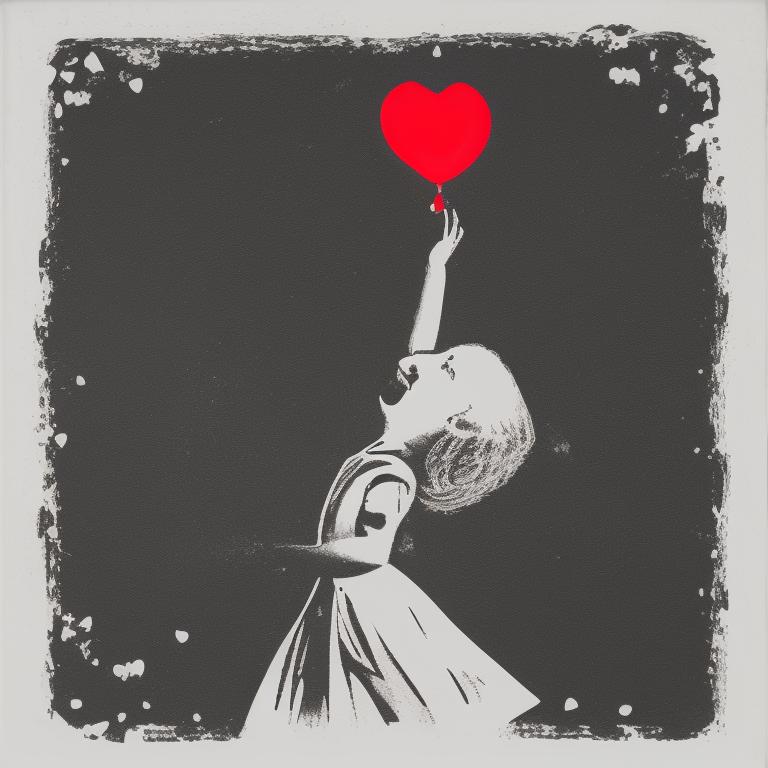} \\
    
    \multicolumn{2}{C{\groupwidth}}{\prompt{A teenage wizard with round glasses}} &
    \multicolumn{2}{C{\groupwidth}}{\prompt{a cartoon beagle in a red dog house}} &
    \multicolumn{2}{C{\groupwidth}}{\prompt{black and white stencil little girl reaching for a red balloon}}
    
    \end{tabular}
    \caption{We compare \modelname-SNC (Ours) to SD2. Our model is less likely to generate iconic characters given suggestive prompts (drawn from \citet{lee2023talkin}).}
    \label{fig:copyright-traps}
    \vspace{-.5cm}
\end{figure}

\subsection{Human evaluation}

While automated quality metrics are useful, given the level of detail and breadth of of the distribution large T2I are intended to generate, there is no substitute for evaluation by human raters. Human pairwise preference ratings for the three 512x512 resolution CommonCanvas models compared to SD2-base can be seen in Figure~\ref{fig:parti-prompts}. 

In this experiment, human raters were shown a prompt (selected randomly from the PartiPrompts prompts set~\citep{yu2022scaling}) along with two generated images in randomized order, one from the reference model (SD2-base) and the other from a CommonCanvas model. Users were asked which generated image they preferred. We report the fraction of the time users selected the image generated by the CommonCanvas model over the corresponding generation from SD2 as the user preference rate for that model. In agreement with our automated quality metrics, we find that the two small CommonCanvas models are less perferred than SD2-base, with preference rates of 37\% for CommonCanvas-SC and 38\% for CommonCanvas-SNC, which we find surprisingly high considering the smaller and synthetic nature of the dataset. For the largest model, CommonCanvas-LNC, we do not measure a statistically significant difference in user preference between this model and SD2-base. While SDXL is a significantly larger model, this finding represents an existential result, showing that we are capable of matching the performance of a model trained on several magnitudes more of data.

\subsection{Benefits and challenges of synthetic captions}

Interestingly, we observe that synthetic captions can enhance the alignment of our model. For instance, the CLIP Score for synthetic captions exceeded that of ground-truth captions as seen in Figure ~\ref{fig:data-variance}. 

We also observed reduced diversity of n-grams in our synthetic captions, a pattern previously noted by~\citet{nguyen2023improving}. 
This effect can be visualized through the decrease in unique trigrams. 

Although we train on Creative-Commons images, it is still possible for an adversarial prompt to produce content that, for example, includes iconic characters. 
In Figure~\ref{fig:copyright-traps}, we subject our model to ambiguous prompts that are suggestive of such characters. Examples include visuals closely resembling Elsa from Frozen, Indiana Jones resembling Harrison Ford, and even a likeness to Harry Potter (Figure~\ref{fig:copyright-traps}). Qualitatively, our model deviated more from these characters than SD2.  

\begin{figure}[t]
    \centering
    \setlength{\groupwidth}{0.31\linewidth}
    \setlength{\itemwidth}{0.5\groupwidth}
    \setlength{\tabcolsep}{0pt}
    \newcolumntype{C}[1]{>{\centering\arraybackslash}p{#1}}
    
    \begin{tabular}{cc@{\hskip 0.05in}cc@{\hskip 0.05in}cc}
    
    Ours & SD2 & Ours & SD2 & Ours & SD2 \\
    \includegraphics[width=\itemwidth]{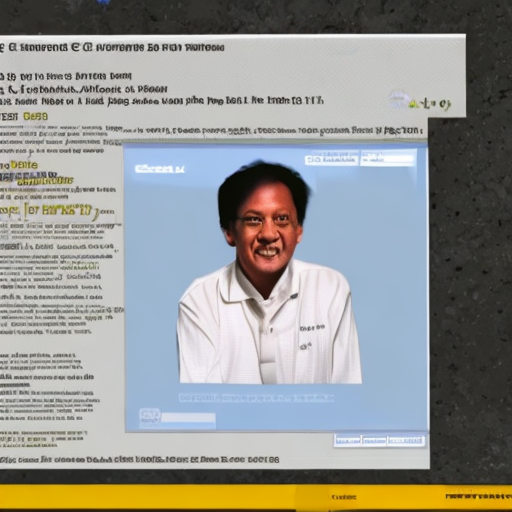} &
    \includegraphics[width=\itemwidth]{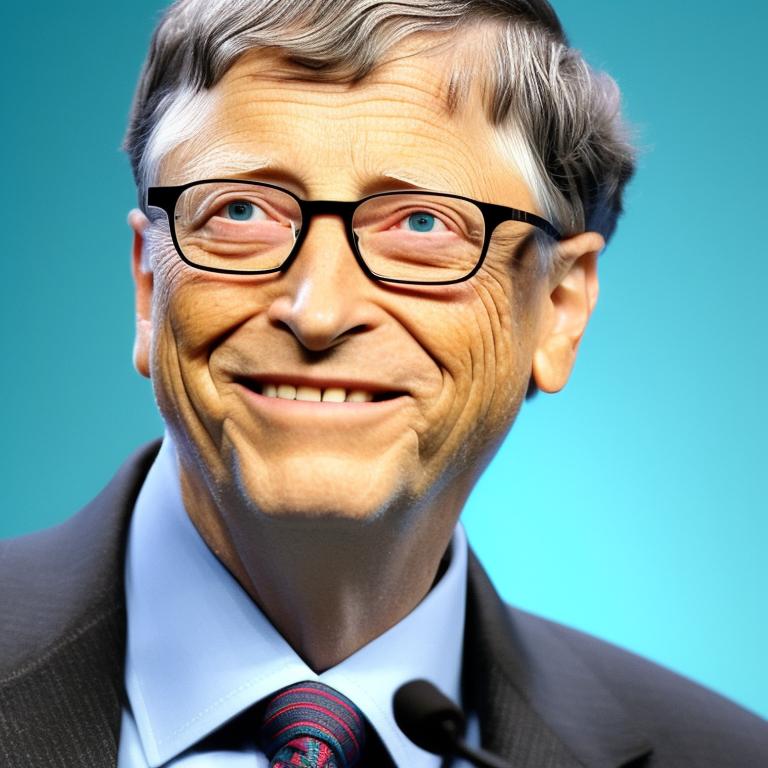} &

    \includegraphics[width=\itemwidth]{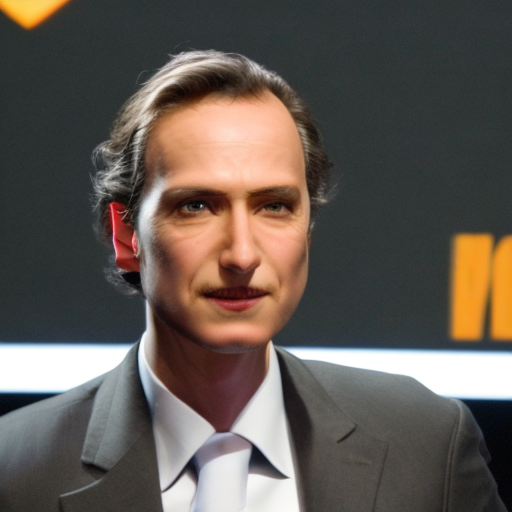} &
    \includegraphics[width=\itemwidth]{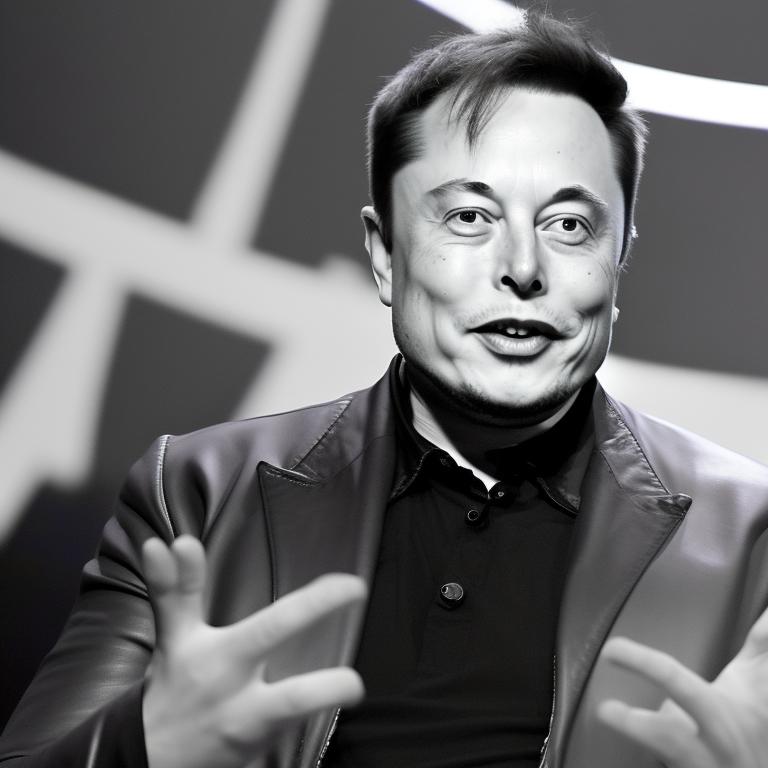} &     \includegraphics[width=\itemwidth]{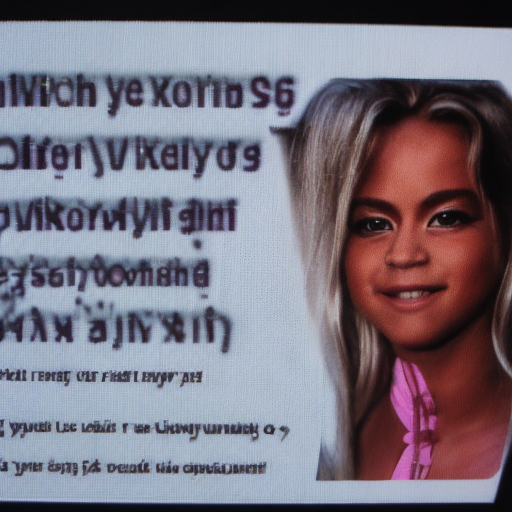} & %
    \includegraphics[width=\itemwidth]{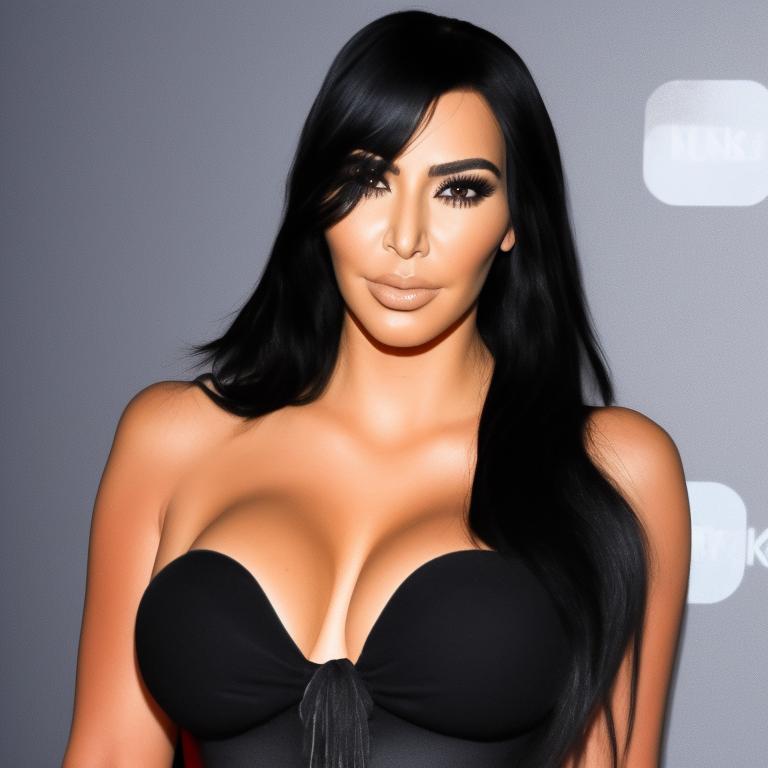} \\

\multicolumn{2}{c}{\prompt{Bill Gates}} & \multicolumn{2}{c}{\prompt{Elon Musk}} &
\multicolumn{2}{c}{\prompt{Kim Kardashian}}

 \\

    \includegraphics[width=\itemwidth]{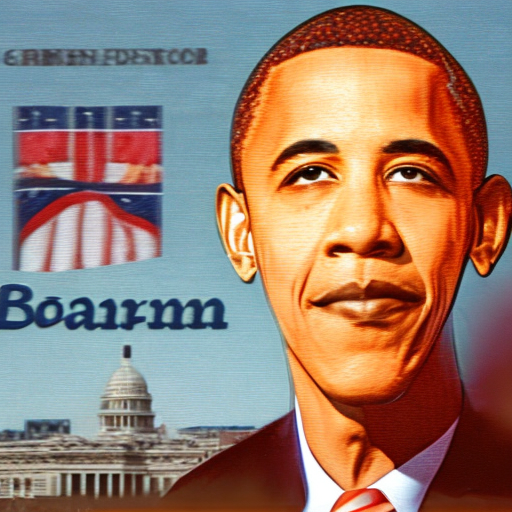} & \includegraphics[width=\itemwidth]{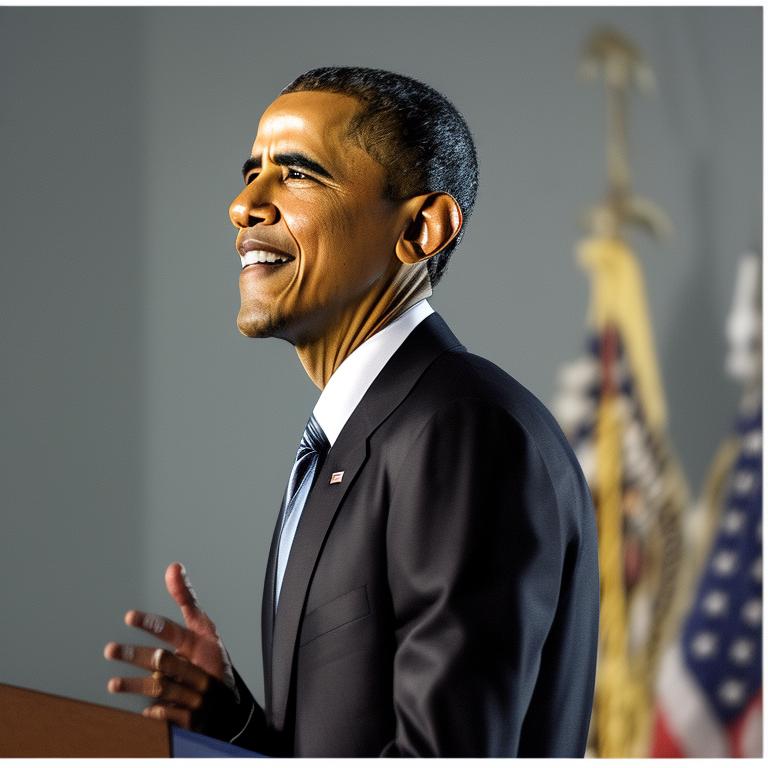} & \includegraphics[width=\itemwidth]{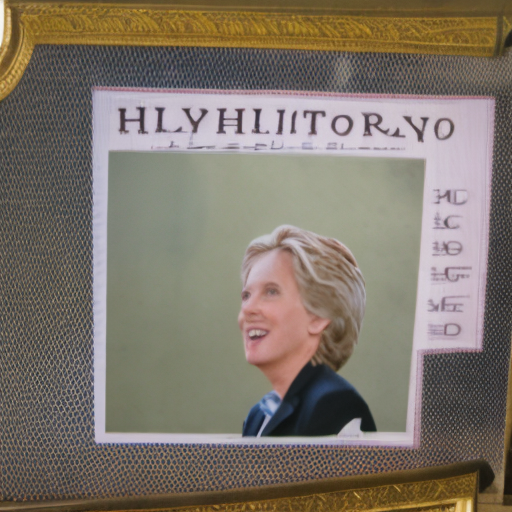} & \includegraphics[width=\itemwidth]{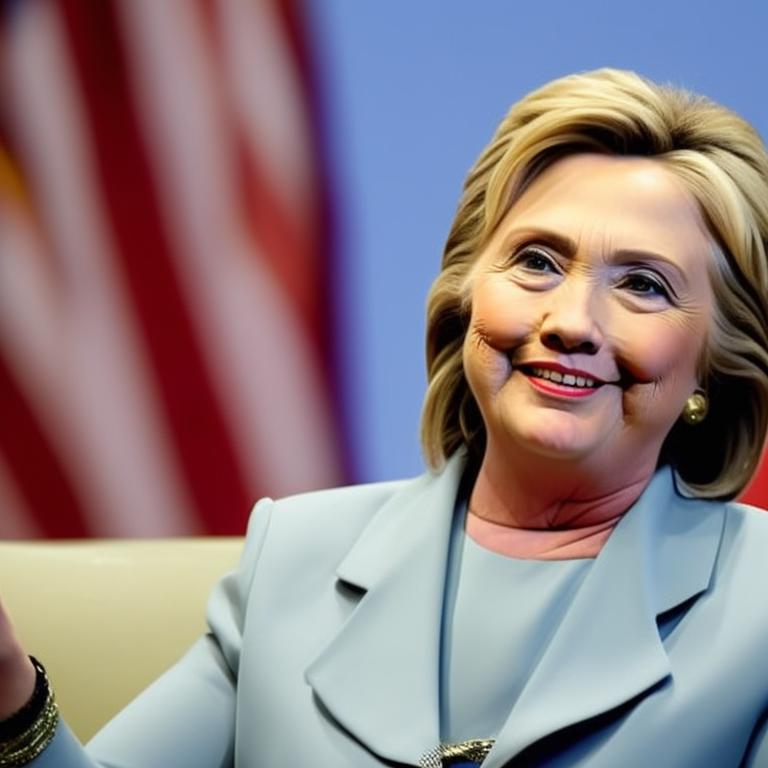} & 
    \includegraphics[width=\itemwidth]{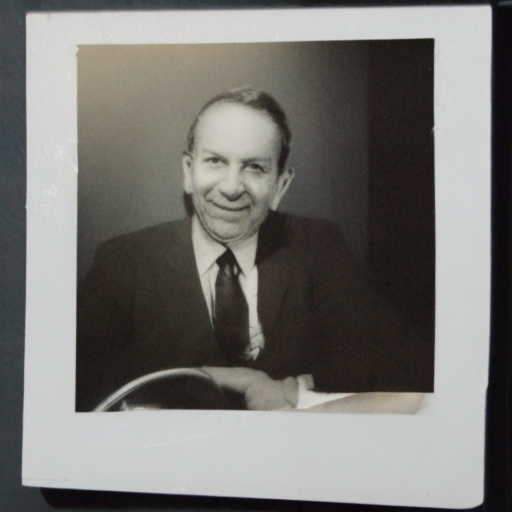} & 
    \includegraphics[width=\itemwidth]{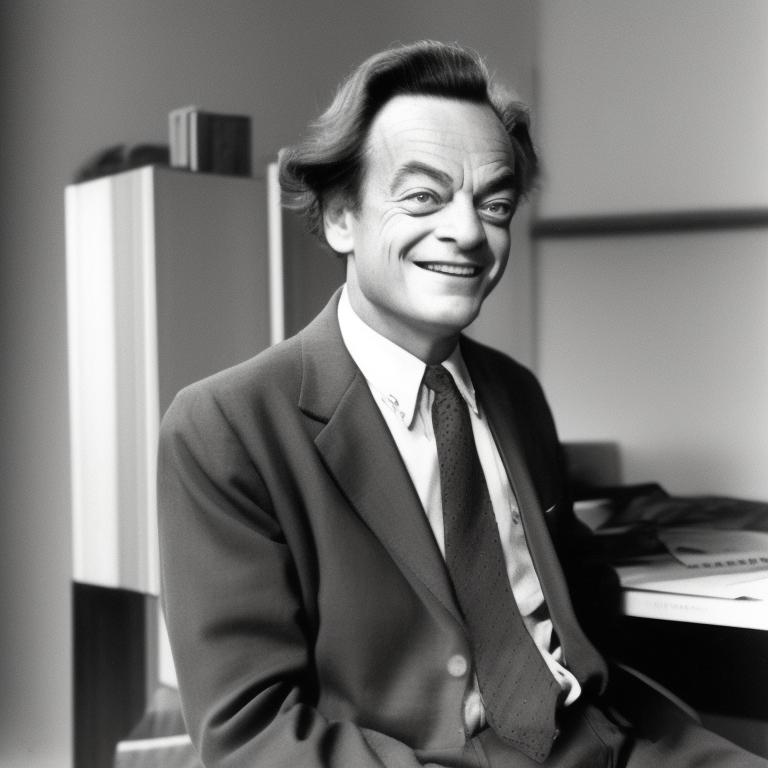} \\
    \multicolumn{2}{c}{\prompt{Barack Obama}} &
    \multicolumn{2}{c}{\prompt{Hillary Clinton}} &
    \multicolumn{2}{c}{\prompt{Richard Feynman}}
    
    \end{tabular}
    \caption{Using \modelname-SNC (Ours) to generate celebrities. Our model is worse at synthesizing individual people than SD2, but is capable of generating some noteworthy public figures.}
    \label{fig:celeb-ids}
    \vspace{-.15in}
\end{figure}


%% file: iclr2023/wandb_charts/conc-caps-subsets/concept-captions-fig.tex
\begin{figure}[t]
    \centering
     \includegraphics[width=0.8\linewidth]
     {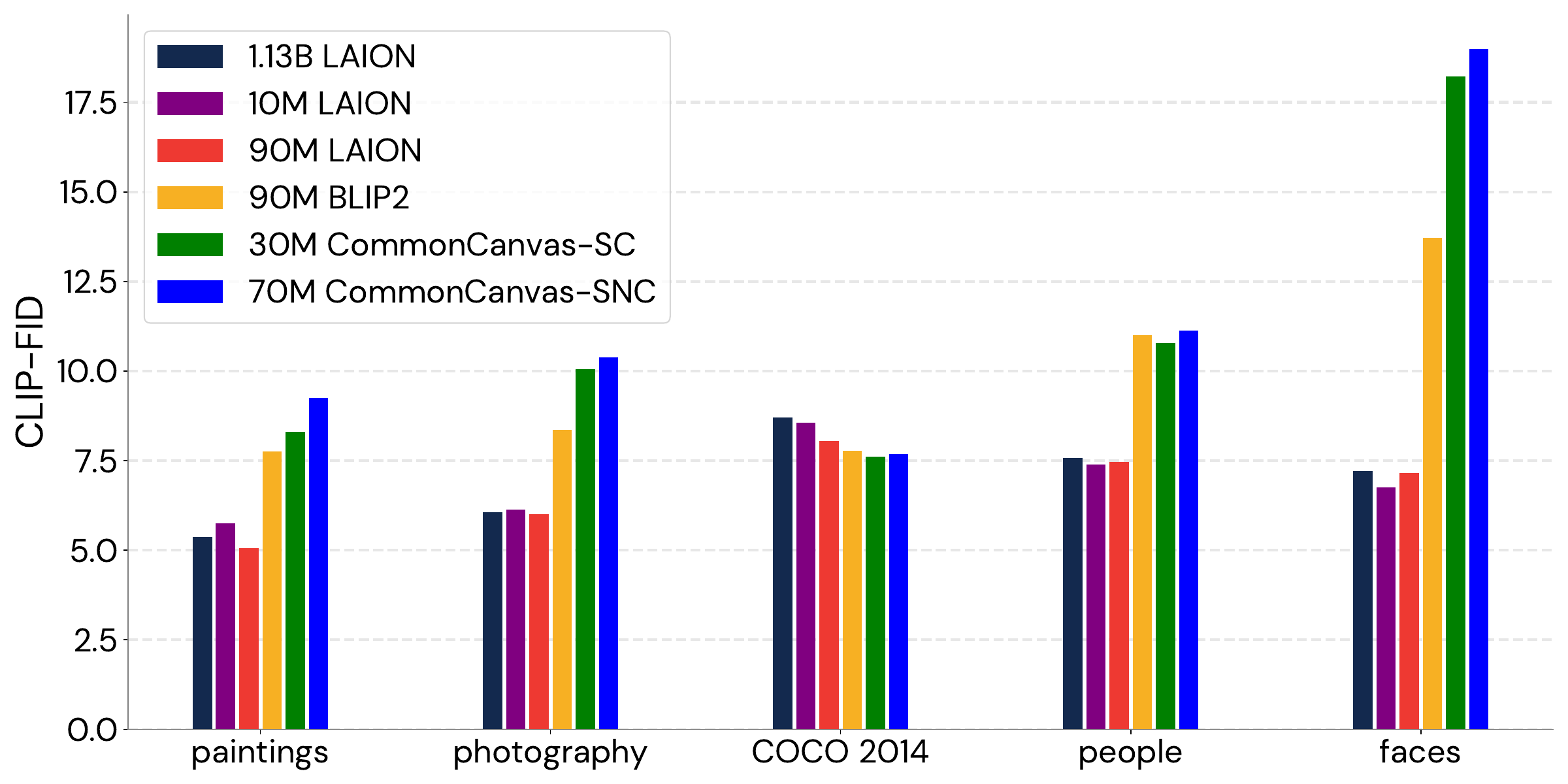}
    \caption{CLIP-FID for different models. We can see domain shift between MS COCO captions and web-scraped conceptual captions. CLIP-FID likely favors SD2, as CLIP is trained on a similar style of text as LAION. This plot only covers the first stage of training at 256x256 resolution. We 
    }
    \label{fig:conceptual-captions}
    \vspace{-.25in}
\end{figure}

%% file: iclr2023/section/70-rw.tex
\vspace{-.2cm}
\section{Discussion and Related Work}\label{sec:discussion}
\vspace{-.2cm}

In this paper, we train the family of \modelname{} text-to-image latent diffusion models on only Creative-Commons images and synthetic captions. 
We discuss the data incompleteness and scarcity issues associated with CC images, and how we address each of these issues in turn.
For data incompleteness, we propose \captionmethod, an intuitive type of transfer learning (Section~\ref{sec:transfer}), which we instantiate with BLIP-2 to produce synthetic captions for CC images --- together, the \datasetname{} dataset (Section~\ref{sec:dataset}). 
With regard to data scarcity, we hypothesize that much less data than what is contained in LAION-2B is necessary to saturate SD2, and that \datasetname{} should be sufficient for training.
To make testing this hypothesis more efficient, we implement a variety of ML-systems optimizations, which achieve a 2.7X speed-up over our SD2 baseline. 
Ultimately, we find that we can train SD2 on $<$3\% of LAION-2B (Section~\ref{sec:mlsys}), which encourages us to train on  \datasetname's commercial (roughly 70 million) and non-commercial (roughly 25 million) examples. 
Our \modelname{} models under-perform in some categories, like faces, but \modelname-LNC demonstrates statistically equivalent performance with SD2 on human evaluation (Section~\ref{sec:experiments}). 

We note that several recent works study copyright.
This work tends to concern text-to-text training data~\citep{min2023silo}, be primarily theoretical~\citep{vyas2023provable, scheffler2022formalizing}, involve ablation studies~\citep{kumari2023ablating}, or only handle verbatim memorization~\citep{carlini2021extracting} through the use of generation-time content filters~\citep{copilot-copy-filter}, which has been shown to be an incomplete solution~\citep{ippolito2023preventing}. 
To the best of our knowledge, no prior open work attempts to train T2I models on only open-licensed data. 


Most prior work on text-caption-dataset creation has focused on extracting caption data from Common Crawl~\citep{gadre2023datacomp,desai2021redcaps,laurencon2023obelics}. 
We instead focus on synthesizing captions directly by using a pre-trained BLIP-2 model.
\cite{nguyen2023improving} demonstrate that existing caption datasets can be improved by using BLIP2 to re-caption low-quality captions in large datasets like Datacomp, but do not focus on creating a new dataset of synthetic captions, as we do here.\looseness=-1 

An issue, which we do not address, is that the YFCC100M data is about a decade old; its CC images are not as current as those in LAION-2B. Given the success of our results, in the future, we plan to augment \datasetname{} with Creative-Commons images from other sources, as well as test larger \modelname{} model architectures. 

%% file: iclr2023/section/80-acknowledgements.tex
\section*{Acknowledgements}

We would like to thank Christopher De Sa  for feedback on earlier drafts of this work. 
A. Feder Cooper is funded by Professor Christopher De Sa's NSF RI-CAREER award 2046760.  This work was also sponsored by Volodymyr Kuleshov's CAREER grant: \#2145577. We also would like to thank Apolinário Passos for helping us host the data + models and for insightful discussions along the way. 

%% file: iclr2023/section/98-appendix.tex
\appendix

\input{iclr2023/section/appendix/15-method}
\input{iclr2023/section/appendix/20-datasets}
\input{iclr2023/section/appendix/30-YFCC-default-captions}
\input{iclr2023/section/appendix/40-training-details}
\input{iclr2023/section/appendix/50-additional-qual-ex}
\input{iclr2023/section/appendix/60-ml-sys}
\input{iclr2023/section/appendix/70-stable-training}

\newpage

%% file: iclr2023/section/appendix/15-method.tex
\section{Additional Details on Data Scarcity Analysis}\label{app:sec;data-scarce} 

\subsection{Hypothesis: Diffusion models are too small}
\label{app:sec;diff-too-small}

A back-of-the-envelope calculation provides some insight on why this is the case. Consider a training dataset consisting of $N$ images with resolution $H\times W$ and $c$ channels. To completely memorize the training data, the model must be capable of storing $c\times H \times W\times N$ numbers. Given a number of trainable parameters $N_p$, it is natural to assume that on average each parameter is capable of storing roughly enough information to reconstruct a single number from the training dataset. Under this assumption, complete memorization is only possible if the size of the training dataset is at or below a critical size $N_c$ ($N\leq N_c$) with $N_c$ given by $N_c=\frac{N_p}{cHW}$. Note that this critical size assumes the data cannot be further compressed, which is obviously not the case for natural images. However, SD2 and SDXL are latent diffusion models, which first use a pretrained encoder to compress images by a factor of $8$ in both $H$ and $W$, and so when we train LDMS like SD2 and SDXL, we are training on data that has been significantly compressed already. 

In our experiments, $c=4$ and $H=W=32$, corresponding to $256\times256$ resolution RGB images in the SD2 and SDXL latent space. The SD2 UNet has $N_p=866\times10^6$ trainable parameters, and SDXL's UNet has $N_p=2567\times10^6$. So we calculate $N_c\approx0.2\times10^6$ for SD2 and $N_c\approx0.6\times10^6$ for CommonCanvas-Large; both of these numbers are several orders of magnitude below the size of our YFCC derived datasets, and so even with significant additional data compression we expect that our \datasetname{} datasets should be sufficient to train both SD2 and SDXL. Additionally, this argument predicts that we should only begin to see significant overfitting in these models for datasets of size $N\sim10^6$. These estimates are resolution dependent, and as image resolution increases we expect that $N_c$ will decrease as more information is provided per image. 

\subsection{Increasing model capacity with CommonCanvas-LNC}
\label{app:sec;largeboi}
We also train a variant of SD2 with more trainable parameters, taking the UNet from SDXL. We refer to this model as CommonCanvas-LNC. We adapt the SDXL UNet architecture to SD2 by changing the cross-attention dimensionality to match that of the SD2 text encoder hidden state dimensionality (1024 for SD2 vs. 2048 for SDXL). SDXL also retrains the VAE component in their model, and we use this improved performance VAE as well. Except for these changes, the architecture is identical to that of SD2.

%% file: iclr2023/section/appendix/20-datasets.tex
\section{Training Dataset Details}\label{app:sec:data}

\subsection{LAION-2B}\label{app:sec:laion}

The fact that LAION is not a stable benchmark can lead to multiple reproducability and security issues. Data poisoning attacks would be difficult to detect at the scale of 2 billion parameters. While this could be mitigated by using hash values of the images, then any time the a site decide to re-encode the image, those images would now need to be excluded from the dataset. Furthermore, targeted data poisoning attacks for diffusion models are no longer just academic conjecture. Last year after the release of Stable Diffusion, a protest was launched on ArtStation that had uses upload images that said ``NoAI'' to taint future training data for generative models after artists felt as though their work had been unfairly used to train the models. With the high degree of link rot, targeted attacks are fairly easy. 
Furthermore, reproduction of the experiments becomes virtually impossible. This means any benchmarks that use copies of LAION as ground truth are are likely using differing subsets of the full dataset.






\subsubsection{Sourcing Creative-Commons images}\label{app:sec:catalog:images}




\begin{table}[h!]
\centering
    \caption{CC licenses in YFCC100M. ND means derivative works are not licensed or the license doesn't allow the user to create derivative works. NC means images cannot be used in commercial contexts. \datasetname-C only contains data from the bottom two (yellow) rows, reflecting images licensed for commercial contexts (i.e., roughly 25 million images). \datasetname-NC contains \datasetname-C, and additionally includes the middle two (blue) rows, reflecting images licensed for non-commercial purposes. We do not include the roughly 30 million images in the top two (pink) rows in \datasetname, as they are non-derivative licenses. We do not train on these images. We do, however, produce BLIP-2 captions for them and release those captions as an evaluation set.}\vspace{.1cm}
    \label{tab:yfcc100m-count-table}
    \footnotesize
    \begin{tabular}{lrr}
    \toprule
    \textbf{CC License} & \textbf{\# Images} & \textbf{\% Captioned} \\\midrule
\rowcolor{palepink}
CC-BY-NC-ND-2.0 
& 25,790,117 & 33.52\%  \\\midrule
\rowcolor{palepink}
CC-BY-ND-2.0 
&  4,827,970 & 30.23\% \\\midrule
\rowcolor{paleblue}
CC-BY-NC-2.0 & 12,468,229 & 31.39\% \\\midrule
\rowcolor{paleblue}
CC-BY-NC-SA-2.0 & 28,314,685 & 31.57\%  \\\midrule
\rowcolor{paleyellow}
CC-BY-SA 2.0
& 9,270,079 &  34.05\% \\\midrule
\rowcolor{paleyellow}
CC-BY 2.0 
& 16,962,338 & 28.96\% \\\bottomrule
    \end{tabular}
\vspace{-.25cm}
\end{table}




\subsubsection{Release and documentation}\label{app:sec:catalog:release}


%% file: iclr2023/section/appendix/30-YFCC-default-captions.tex
\section{YFCC Example Images}
\begin{table}[h]
    \centering
    \setlength{\itemwidth}{0.3\linewidth}
    \caption{Randomly sampled images from the YFCC~\citep{thomee2016yfcc100m} training set. Our synthetic BLIP2 captions are also provided below. }
    \newcolumntype{C}[1]{>{\centering\arraybackslash}p{#1}}
    
    \begin{tabular}{C{\itemwidth} C{\itemwidth} C{\itemwidth}}
 \makebox[\itemwidth]{%
                     \includegraphics[width=\itemwidth]{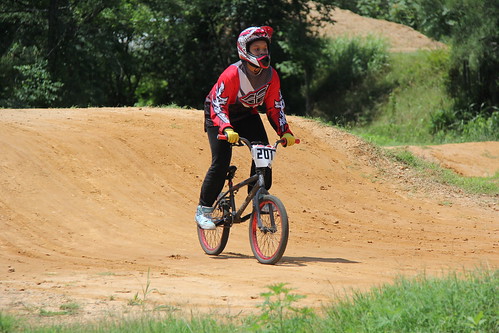}
                    }
         & \includegraphics[width=\itemwidth]{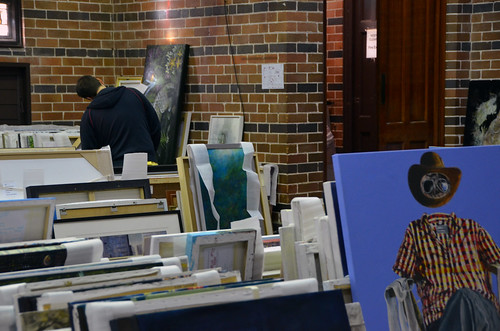}& \includegraphics[width=\itemwidth]{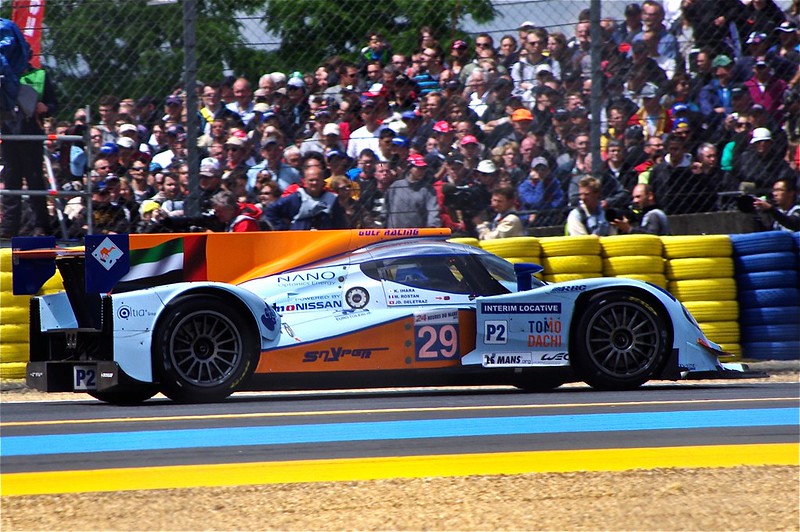} \\ 
         a person riding a bike on a dirt road & a paintings on the wall & an orange and blue race car driving on a track
    \end{tabular}
    
    \label{tab:yfcc-examples}
\end{table}
\input{iclr2023/tables/top10-flickr-captions}

%% file: iclr2023/tables/top10-flickr-captions.tex

\begin{table}[ht]
\caption{Top 10 highest frequency captions in the YFCC dataset. The most common captions are not user generated and are not very descriptive of the corresponding image.
\vspace{.2cm}
}
\centering
\begin{tabular}{|p{0.7\linewidth}|c|}
\hline
\textbf{YFCC Original Caption} & \textbf{Count} \\
\hline
OLYMPUS+DIGITAL+CAMERA & 184889 \\
SONY+DSC & 123128 \\
Exif\_JPEG\_PICTURE & 104480 \\
Barclays+Center+Arena\%0AAtlantic+Yards\%0A6th+and+Atlantic+A
& 68832 \\
Olympus+digital+camera & 54805 \\
Effortlessly+uploaded+by \href{http://www.eye.fi}{Eye-Fi} & 48388 \\
. & 43227 \\
-+Camera+phone+upload+powered+by \href{http://www.shozu.com/?utm\_source=upload&utm\_medium=graphic&utm\_campaign=upload\_graphic}{ShoZu} & 38856 \\
Sony+dsc & 32709 \\
Photo+by \href{http://twitter.com/Kmeron}{@Kmeron} \href{http://www.facebook.com/musicfromthepit}{|Facebook page is this way|} & 23754 \\
\hline
\end{tabular}

\end{table}

\begin{table}[]
    \centering
        \caption{Number of usable captions from OpenAI's YFCC14M dataset~\citep{radford2021clip}. This table is actually a subset from~\ref{tab:yfcc100m-count-table} for which either the user description or image title were deemed usable. These figures provide an estimate on how many images in each category are actually potentially usable as captions.}

\input{iclr2023/tables/openai-license-count}

    \label{tab:openai-usable}
\end{table}

%% file: iclr2023/tables/openai-license-count.tex
\begin{tabular}{lr}
\hline
                              License Name &    count \\
\hline
    CC-BY 2.0 &  2448002 \\
    CC-BY-ND 2.0 &   682273 \\
    CC-BY-NC  2.0 &  1925854 \\
    CC-BY-NC-ND  2.0 &  4058817 \\
    CC-BY-NC-SA  2.0 &  4146113 \\
   CC-BY-SA 2.0 &  1568336 \\
\hline
\end{tabular}

%% file: iclr2023/section/appendix/40-training-details.tex
\textbf{Model Architecture}

We follow the model architecture and training recipe of Stable Diffusion 2 as closely as we can to best reproduce the model for CC-Small. The model has an identical number of params and structure as the original model. In fact, we can even load SD2's model weights into our framework due to the identical architecture and naming scheme. We are able to achieve virtually identical performance with SD2 in a much shorter training time with less data. We use the same VAE, tokenizers, and UNet archicture as SD2 except for reducing the precision of the normalization layers.

Our CC-Large model takes SD2's model and replaces the UNet with the SDXL architecture~\citep{podell2023sdxl}. Like CC-Small, we also replace the normalization layers with their low-precision version. The replacement of all the normalization layers is handled automatically by MosaicML's Composer library~\citep{mosaicml2022composer}. We perform all dataloading through MosaicML's streaming library~\citep{mosaicml2022streaming}.

%% file: iclr2023/section/appendix/50-additional-qual-ex.tex
\begin{figure}
    \centering 
    \setlength{\tabcolsep}{1pt}
    \setlength{\itemwidth}{0.2\linewidth}
    \newcolumntype{M}[1]{>{\centering\arraybackslash}m{#1}}
    \begin{tabular}{M{\itemwidth}M{\itemwidth}M{\itemwidth}M{\itemwidth}M{\itemwidth}}
      Prompt & SD2 & \modelname-SC & \modelname-SNC & \modelname-LNC \\

    a 3D CAD model of an airplane &
    \includegraphics[width=\itemwidth]{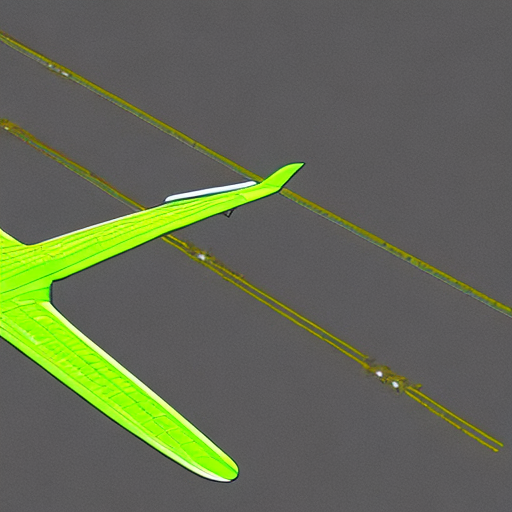}  &
    \includegraphics[width=\itemwidth]{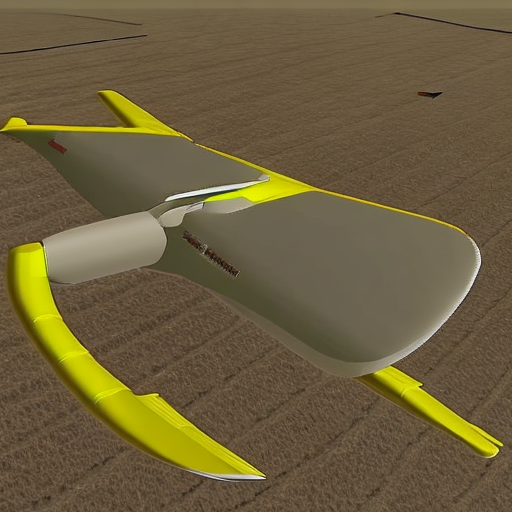} &
    \includegraphics[width=\itemwidth]{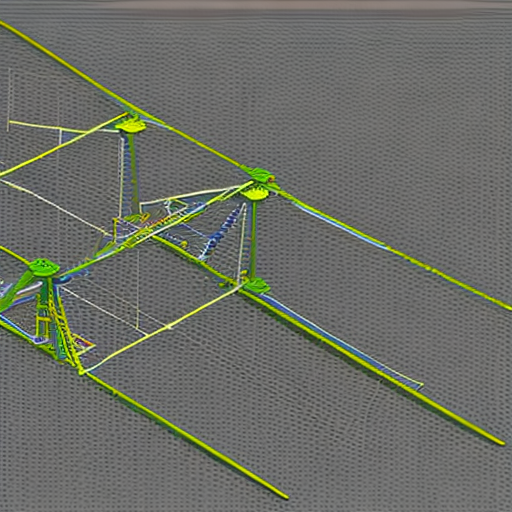} &     
    \includegraphics[width=\itemwidth]{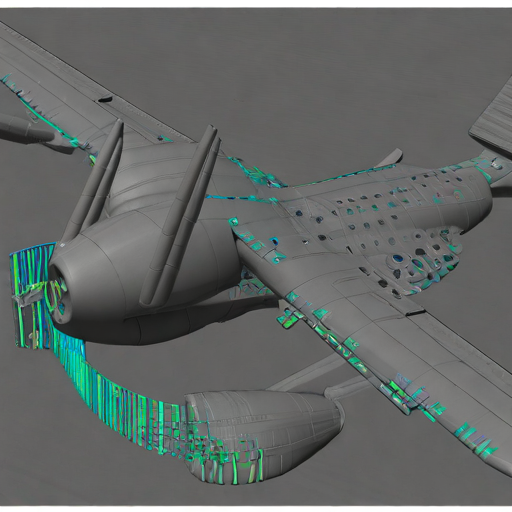} \\

    a bear and a fox in the forest &
    \includegraphics[width=\itemwidth]{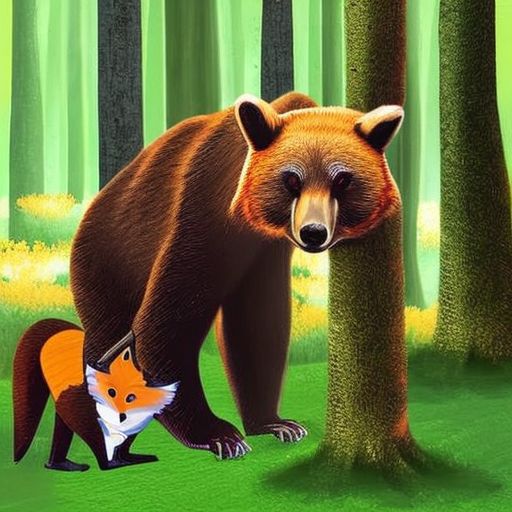}  &
    \includegraphics[width=\itemwidth]{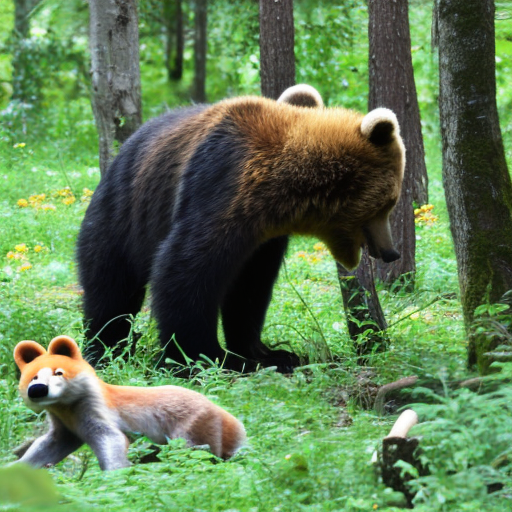} &
    \includegraphics[width=\itemwidth]{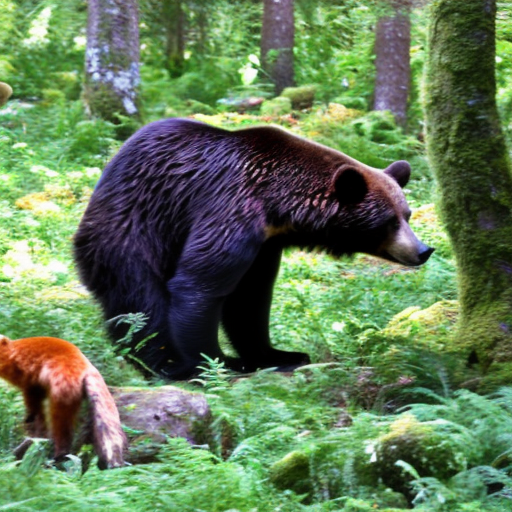} &     
    \includegraphics[width=\itemwidth]{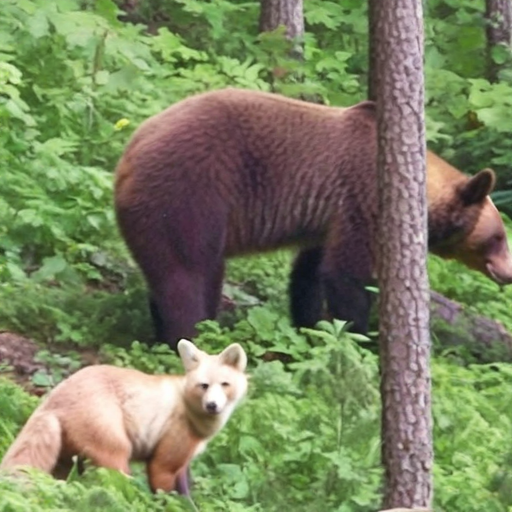} \\


    a klein bottle &
    \includegraphics[width=\itemwidth]{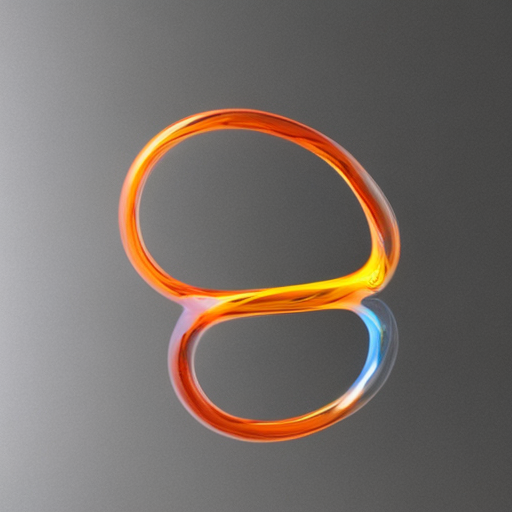}  &
    \includegraphics[width=\itemwidth]{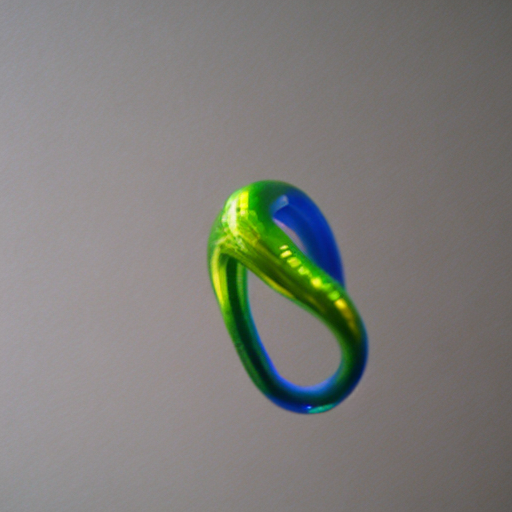} &
    \includegraphics[width=\itemwidth]{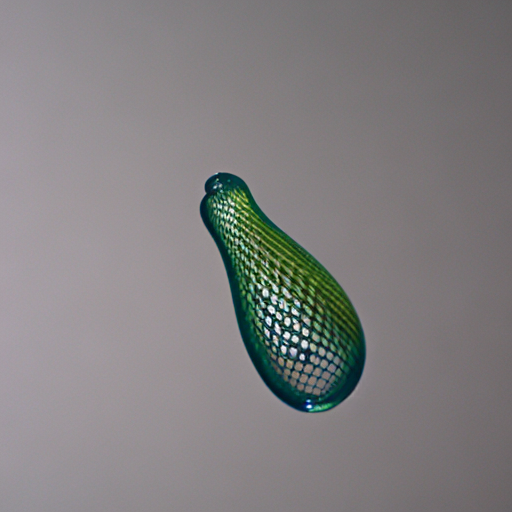} &     
    \includegraphics[width=\itemwidth]{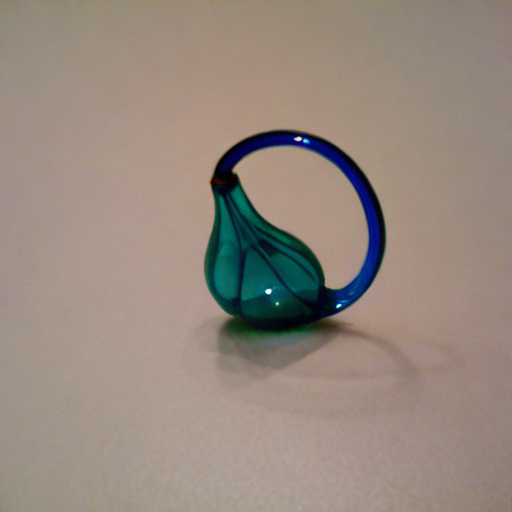} \\

    a partially cut birthday cake with pink and blue frosting &
    \includegraphics[width=\itemwidth]{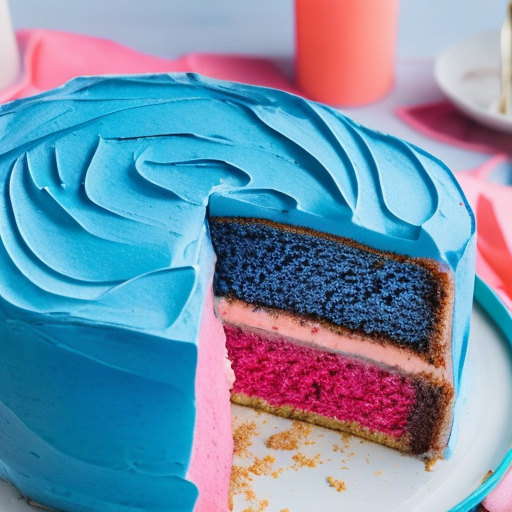}  &
    \includegraphics[width=\itemwidth]{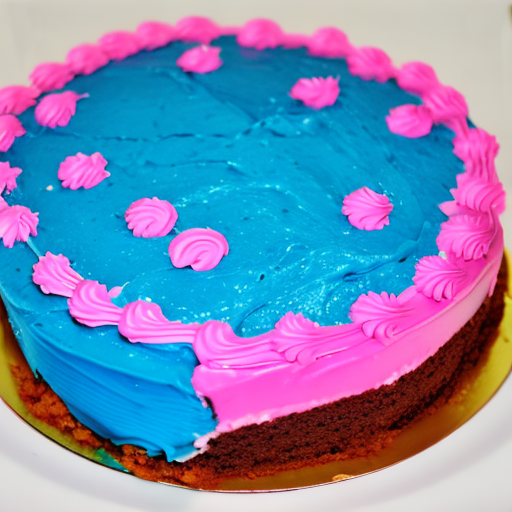} &
    \includegraphics[width=\itemwidth]{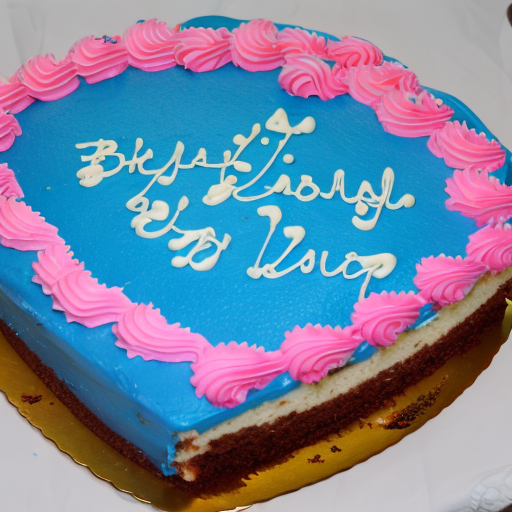} &     
    \includegraphics[width=\itemwidth]{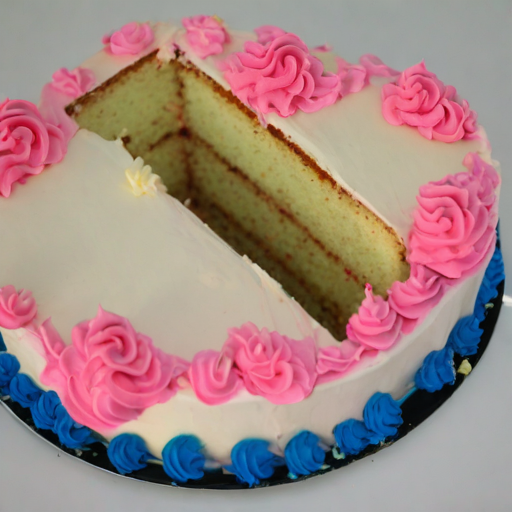} \\


    two hummingbirds and a squirrel in a bird bath &
    \includegraphics[width=\itemwidth]{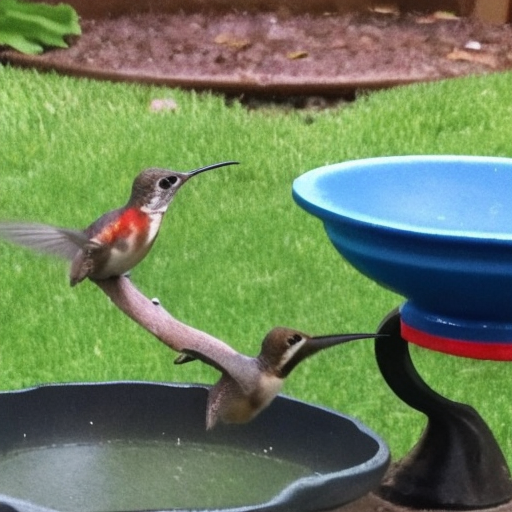}  &
    \includegraphics[width=\itemwidth]{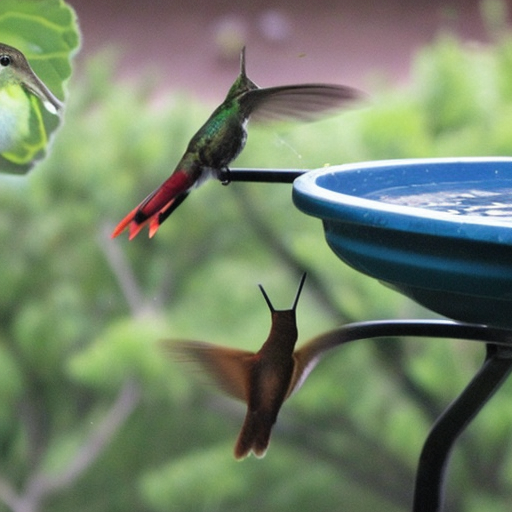} &
    \includegraphics[width=\itemwidth]{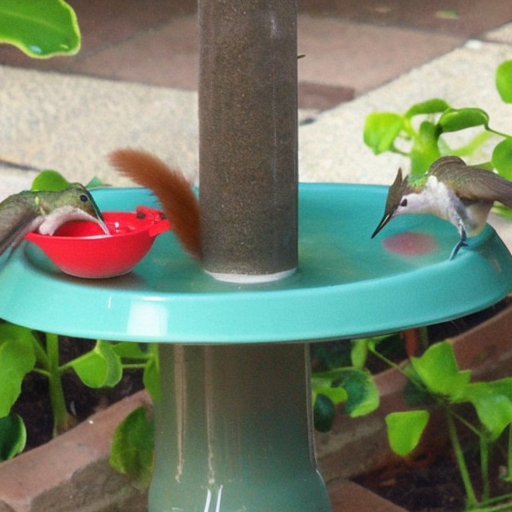} &     
    \includegraphics[width=\itemwidth]{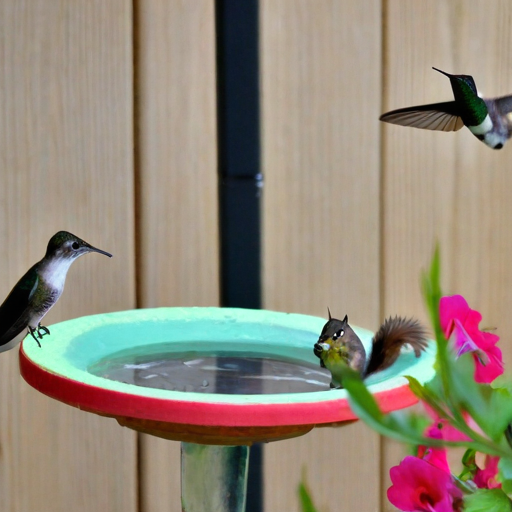} \\
    
    \end{tabular}

    \caption{Additional qualitative examples comparing SD2 to our model trained on the commerical split (\modelname-SC), non-commerical split (\modelname-SNC), and the larger UNet model trained on the non-commercial (\modelname-LNC).}
    \label{fig:additional-qual-exs}
\end{figure}

\begin{figure}
    \centering
    \setlength{\tabcolsep}{1pt}
    \setlength{\itemwidth}{0.2\linewidth}
    \begin{tabular}{c c c c c}
        \includegraphics[width=\itemwidth]{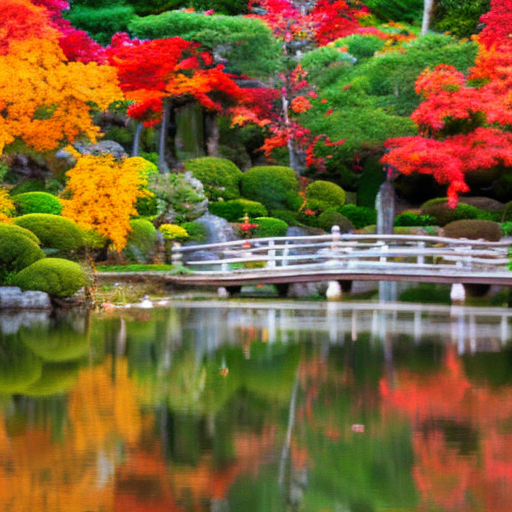}  & 
        \includegraphics[width=\itemwidth]{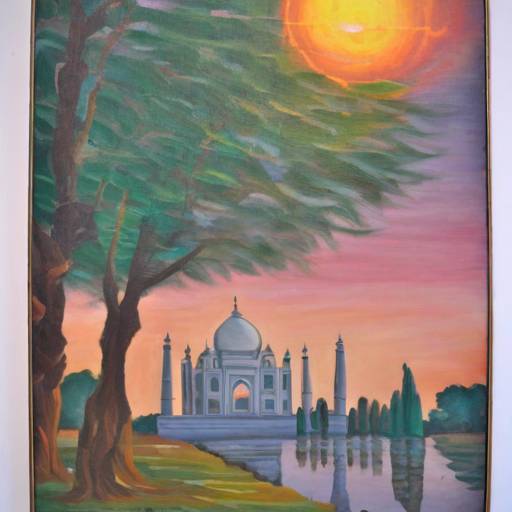}  &         &         \includegraphics[width=\itemwidth]{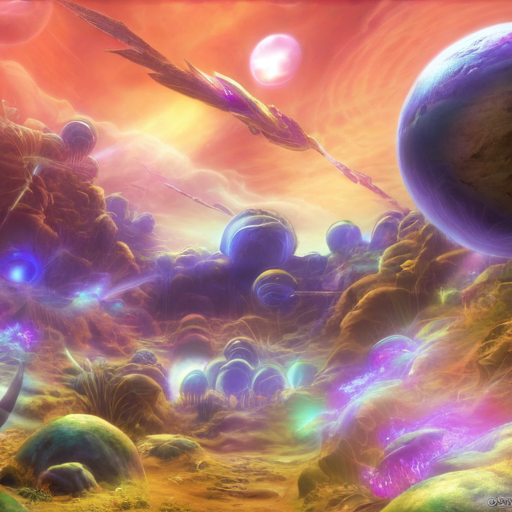}  &         \includegraphics[width=\itemwidth]{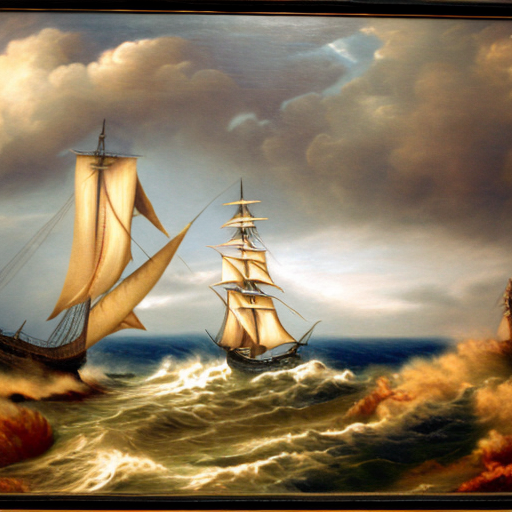}  
        \\ 
    \end{tabular}
    \caption{Additional qualitative examples of our \modelname{} models.}
    \label{fig:enter-label}
\end{figure}

\begin{figure}
    \centering 
    \setlength{\tabcolsep}{1pt}
    \setlength{\itemwidth}{0.2\linewidth}
    \newcolumntype{M}[1]{>{\centering\arraybackslash}m{#1}}
    \begin{tabular}{M{\itemwidth}M{\itemwidth}M{\itemwidth}M{\itemwidth}M{\itemwidth}}
      Input for BLIP2 & BLIP2 Caption & SD2 & \modelname-SNC & \modelname-SC \\

        \makebox[\itemwidth]{%
            \includegraphics[width=\itemwidth,height=\itemwidth,keepaspectratio]{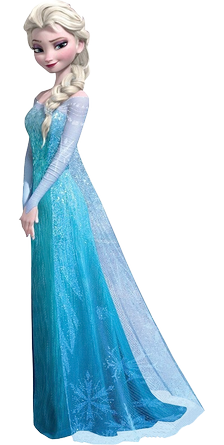}%
        }%
    &
          an image of elsa from frozen  &
    \includegraphics[width=\itemwidth]{iclr2023/figs/prompt-roundtrips/elsa-from-frozen/SD2.png} &
    \includegraphics[width=\itemwidth]{iclr2023/figs/prompt-roundtrips/elsa-from-frozen/YFCC-NC.png} &     \includegraphics[width=\itemwidth]{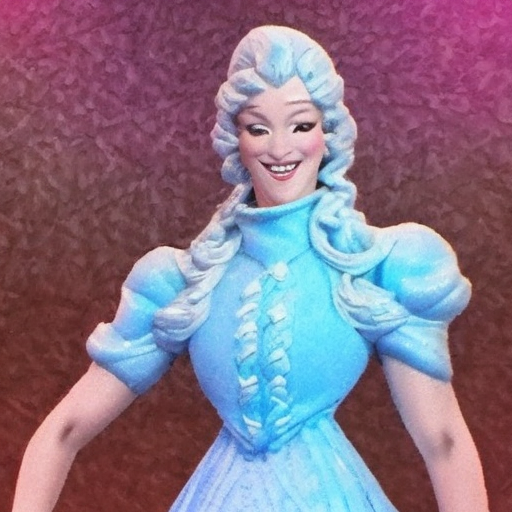} \\

    \makebox[\itemwidth]{%
            
    \includegraphics[width=\itemwidth,height=\itemwidth,keepaspectratio]{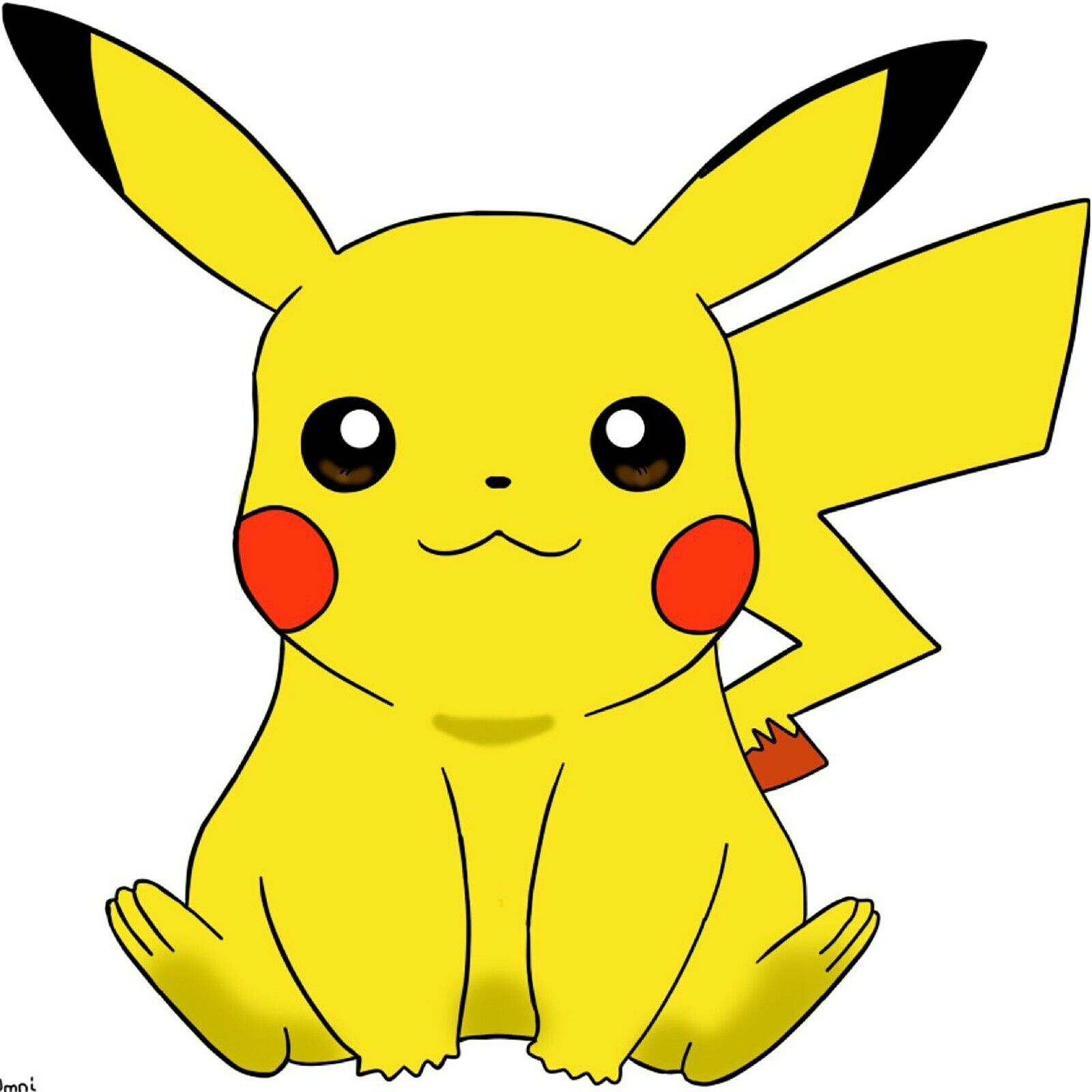}  
    } & 
        pikachu pikachu pikachu pikachu pikachu pikachu pikachu pikachu pikachu pikachu &
    \includegraphics[width=\itemwidth]{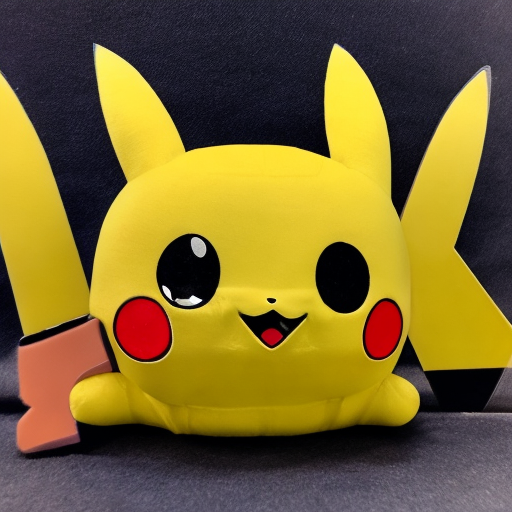} &
    \includegraphics[width=\itemwidth]{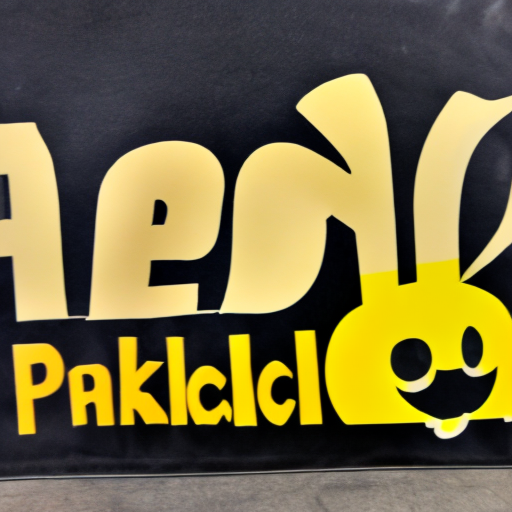} &     \includegraphics[width=\itemwidth]{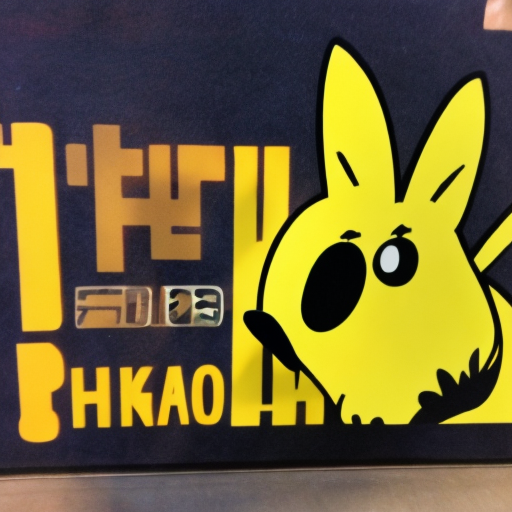} \\
        \makebox[\itemwidth]{%
            
    \includegraphics[width=\itemwidth,height=\itemwidth,keepaspectratio]{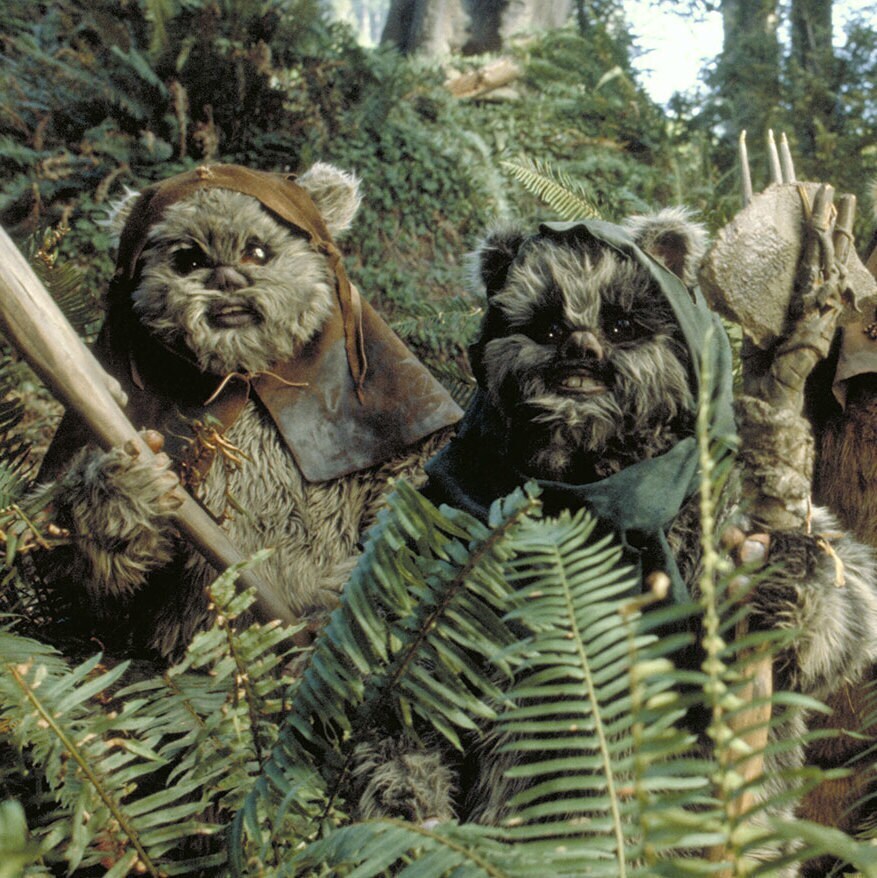}  
    } & 
    three characters dressed like bears, standing in the forest &
    \includegraphics[width=\itemwidth]{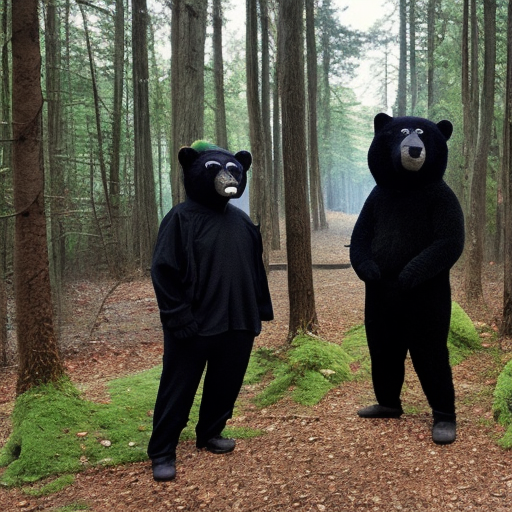} &
    \includegraphics[width=\itemwidth]{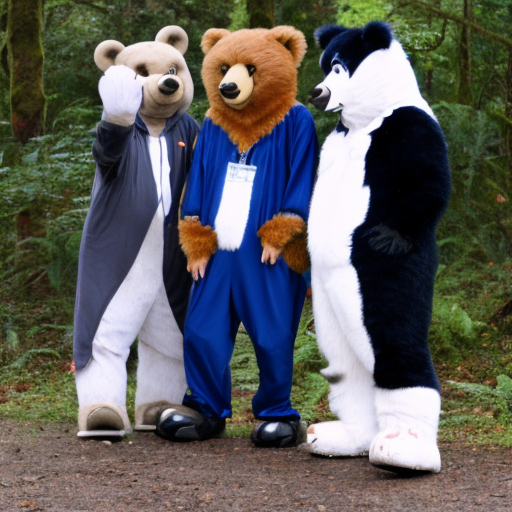} &     
    \includegraphics[width=\itemwidth]{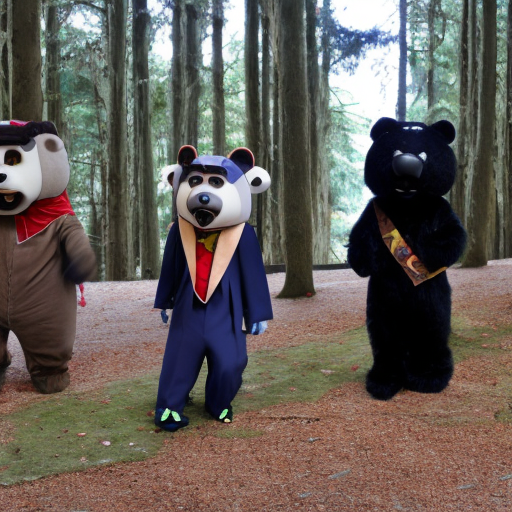} \\
    \end{tabular}

    \caption{Additional qualitative examples comparing our \modelname{} models to SD2, given synthetic BLIP2 captions as prompts. While not perfect, our models are better at avoiding generating potentially problematic data. }
    
    \label{fig:blip-roundtrip}
\end{figure}

%% file: iclr2023/section/appendix/60-ml-sys.tex
\section{Additional Details on Efficiency Optimizations}\label{app:sec;mlsys}

\input{iclr2023/tables/speedup-table}


In this section we provide additional details on the optimizations we implemented to achieve SD2 training speedups. We also report the approximate cost of training our implementation of SD2 on various hardware configurations in Table~\ref{table:speeduptable}.

\custompar{Flash Attention} Cross attention operations are a very expensive part of training that occurs in dozens of layers in diffusion model UNets~\citep{rombach2022diffusion}. Flash Attention is an efficient implementation that is optimized to work well with reduced precision and GPU hardware~\citep{dao2022flashattention}, which was implemented using the XFormers library~\citep{xFormers2022}, allowing us to save compute and memory usage.

\custompar{Precomputing latents} 
Each forward pass of SD2 requires computing a latent representation of the input image, as well as transforming the caption into a text embedding.
Instead of computing the latents for each example during training, we can precompute latents for the entire dataset, amortizing the cost.
Doing so speeds up training of the model, especially at lower resolutions, in exchange for a one-time fixed cost of precomputing all the latents over 1 epoch. 

\custompar{Reduced-precision GroupNorm and LayerNorm}
Most layers in SD2 are implemented in \textsf{float16} precision, but GroupNorm and LayerNorm are implemented in \textsf{float32}, in part because it was assumed to be necessary for training stability. The resulting, frequent upcasting 
causes a major bottleneck in training speed. Recent work shows that it is safe to implement LayerNorm using \textsf{float16} precision~\citep{portes2023mosaicbert}, and we found the same to be true of GroupNorm.
We thus cast all GroupNorm and LayerNorm operators to \textsf{float16} and are able to further reduce total memory consumption and accelerate training.

\custompar{Fully-Sharded Data Parallelism (FSDP)} 
FSDP is a variant of data-parallel training that shards the models parameters, gradients and optimizer state across multiple devices. When training data batches  do not fit into memory, we do several forward and backward passes on smaller microbatches, followed by a single gradient update. At GPU scale, there may only be a single microbatch, so the time for the gradient update can become a significant bottleneck. In standard data distributed training, each GPU communicates all its gradients to every other GPU, and then each GPU updates its local copy of the model. Instead, we use a different paradigm inspired by~\citep{xu2020automatic} where each GPU only gets the gradients and updates the weights for a small part of the model before sending the updated weights for that part of the model to all of the other GPUs. By dividing the update step across all the GPUs, we can ensure that the amount of work per GPU decreases as we increase the number of GPUs, helping us achieve linear scaling. To tackle this problem, we use PyTorch's experimental support for Fully Sharded Data Parallelism (FSDP), specifically, FSDP’s SHARD\_GRAD\_OP mode.  

\custompar{Scheduled Exponential Moving Average (EMA)}
SD2 uses EMA, which maintains an exponential moving average of the weights at every gradient update for the entire training period. This can be slow due to the memory operations required to read and write all the weights at every step. Since the old weights are decayed by a factor of 0.9999 at every batch, the early iterations of training only contribute minimally to the final average. We decide to only apply EMA for the final 50K steps (about 3.5\% of the training period), and are able to avoid adding overhead and still achieve a nearly equivalent EMA model.

\section{\capcaptionmethod: A Transfer Learning-based Image-captioning Method}
\vspace{-.2cm}

Our solution for handling the lack of captions in CC images is called \emph{\captionmethod{}}, a type of transfer learning (Figure~\ref{fig:telephoning}). 
\capcaptionmethod{} assumes the existence of a large labeled dataset $\mathcal{D}_1 = \{(x^{(i)}, y^{(i)})\}_{i=1}^n$, consisting of pairs of high-dimensional $x^{(i)}$ (e.g., images, audio) that map to a compact, structured label $y^{(i)}$ (e.g., caption, audio transcript). \capcaptionmethod{} trains a forward model $q(y | x)$ on $\mathcal{D}_1$ to learn the mapping of $y$ given $x$ via maximum likelihood learning $\max_{q \in \mathcal{Q}} \sum_{i=1}^n \log q(y^{(i)} | x^{(i)})$. It then uses $q$ as training signal for a reverse model $p(x|y)$ trained on a {\em separate} dataset $\mathcal{D}_2 = \{x^{(i)}\}_{i=1}^m$ by maximizing $\sum_{i=1}^m \mathbb{E}_{y \sim q(y|x^{(i)})} [\log p(x^{(i)} | y^{(i)})]$, the likelihood of the data $\mathcal{D}_2$ and the predicted label $y$ under $q$. This forms a type of knowledge transfer from the forward labeling task defined by $\mathcal{D}_1$ to the reverse task of inverting $x$ from $y$ on a separate $\mathcal{D}_2$.

While \captionmethod{} can be viewed as a type of synthetic labeling, it becomes particularly interesting when $x$ is a type of protected modality (e.g., a copyrighted image), while $y$ is a compact representation of $x$ that does not encode sensitive aspects of $y$ (e.g., a generic caption).
Effectively, \captionmethod{} performs a type of ``lossy compression'' or ``distillation" from a high-dimensional  or information-rich $x$ (e.g., an image of Snoopy) to a low-dimensional or information-poor $y$ that loses the sensitive content in $x$ (e.g., the visual characteristics of Snoopy).
Because this compression step is ``lossy'', a reconstruction $x'$ of $x$ from $p(x|y)$ via $y$ often does not remotely resemble the original input, just like in a game of telephone~\citep{telephone}. 
We derive the term \captionmethod{} from the above intuition, and employ it as useful shorthand to denote instances of transfer learning that solve data-scarcity problems in multimodal generative modeling.  

\paragraph{\capcaptionmethod{} for text-to-image modeling.}
In this work, we apply \captionmethod{} to the image and text domains, where CC images are the high-dimensional inputs $x$, and we use a pre-trained BLIP-2 model~\cite{li2023blip2} for ``lossy compression'' to short-text captions $y$ (Figure~\ref{fig:telephoning}a).
Together, these CC-image-caption pairs comprise  the \datasetname{} dataset, which we use to train our \modelname{} T2I models (Figure~\ref{fig:telephoning}b). 
Even though BLIP-2 was pre-trained on LAION-400M~\citep{laion400}, \datasetname{} and \modelname{} never have direct access to LAION-400M or, importantly, anything that is similar to the images that BLIP-2 was trained on. 
Instead, we only have access to the mapping in the model, which, given an image input, produces lossy output text that inherently does not literally resemble its image counterpart (Figure~\ref{fig:telephoning}c).\footnote{We draw on the example of Snoopy from~\cite{sag2023safety}. Figure~\ref{fig:telephoning}'s Snoopy is CC-licensed~\citep{snoopypic}.}   

%% file: iclr2023/tables/speedup-table.tex
\begin{table}
\centering
\caption{Performance (throughput) and approximate cost of training SD2 UNet with our optimizations. Depending on the number of GPUs used, the cost to train the same models without these optimizations range from \$90,000-\$140,000}
\label{tab:unet-performance-cost}
\resizebox{\textwidth}{!}{%
\small 
\begin{tabular}{|c|c|c|c|c|c|}
\hline
\textbf{Number of A100s} & \textbf{256x256 (img/s)} & \textbf{512x512 (img/s)} & \textbf{512x512 with EMA (img/s)} & \textbf{Days to Train} & \textbf{Cost (\$)} \\
\hline

8  & 1100 & 290  & 290  & 101.04 & \$38,800.00 \\
\hline

16 & 2180 & 585  & 580  & 50.29  & \$38,630.00 \\
\hline

32 & 4080 & 1195 & 1160 & 25.01  & \$38,420.00 \\
\hline

64 & 8530 & 2340 & 2220 & 12.63  & \$38,800.00 \\
\hline

128& 11600& 4590 & 3927 & 6.79  & \$41,710.00 \\
\hline
\end{tabular}%
}
\label{table:speeduptable}
\end{table}


%% file: iclr2023/section/appendix/70-stable-training.tex
\begin{figure}
    \centering
    \includegraphics[width=\linewidth]{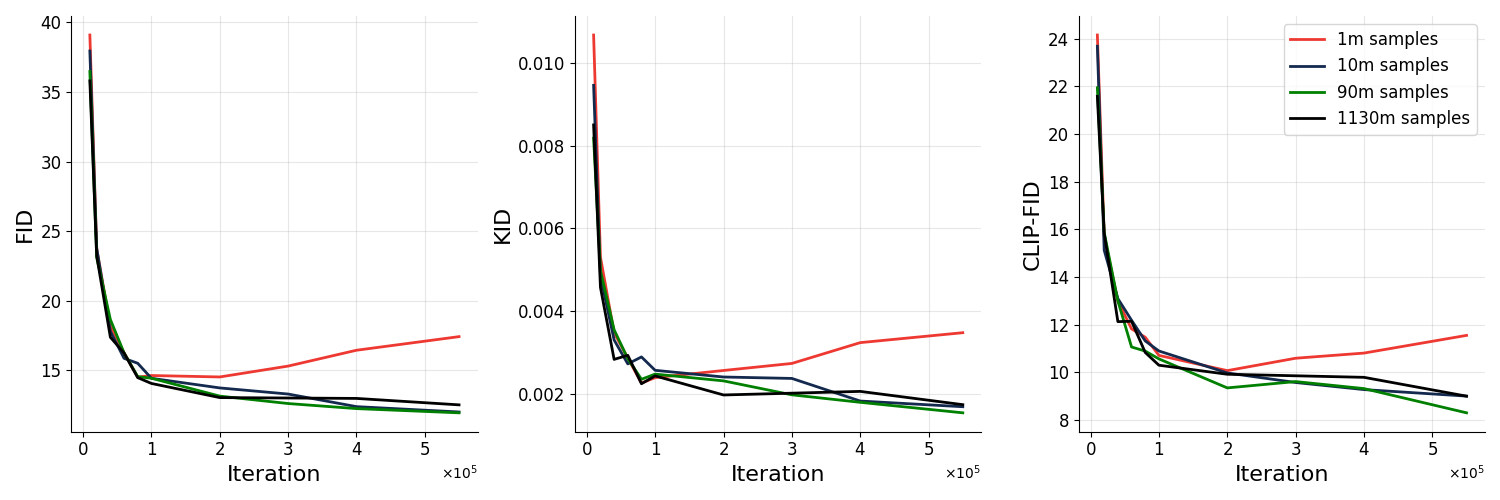}
    \caption{How does reducing the amount of training data affect the training dynamics? We find a noticeable improvement drop when training with less than 10 million samples.}
    \label{fig:eval-over-time-less-data}
\end{figure}